\documentclass[journal]{IEEEtran}
\usepackage{amsmath,amsfonts}
\usepackage{algorithmic}
\usepackage{algorithm}
\usepackage{array}
\usepackage[caption=false,font=scriptsize,labelfont=sf,textfont=sf]{subfig}
\usepackage{textcomp}
\usepackage{stfloats}
\usepackage{url}
\usepackage{verbatim}
\usepackage{graphicx}
\usepackage{cite}
\usepackage{siunitx}
\usepackage{nicefrac}
\usepackage{amsthm}
\usepackage[dvipsnames]{xcolor}
\usepackage{hyperref}
\usepackage{multirow}
\usepackage{multicol}

\newtheorem{definition}{Definition}
\newtheorem{proposition}{Proposition}

\usepackage{tikz}
\usetikzlibrary{shapes,arrows}
\tikzstyle{block} = [rectangle, fill, fill=teal!10, 
    text width=12em, text centered, minimum height=2em]
\tikzstyle{empty} = [rectangle, fill, fill=white, 
     text centered]
\tikzstyle{line} = [draw, -latex']
\usetikzlibrary{calc}

\newcommand{\normtwo}[1]{\left\lVert#1\right\rVert_2}

\newcommand{\removethis}[1]{}

\newcommand{\mR}{\mathcal{R}} 
\newcommand{\mRx}[1]{\mathcal{R_{#1}}}
\newcommand{\mO}{\mathcal{O}}
\newcommand{\mC}{\mathcal{C}}
\newcommand{\mH}{\mathcal{H}}

\newcommand{\mD}{\mathcal{D}}

\newcommand{\mS}{\mathcal{S}}
\newcommand{\mP}{\mathcal{P}}

\newcommand{\mQ}{\mathcal{Q}}

\newcommand{\mJ}{\mathcal{J}}

\newcommand{\mM}{\mathcal{M}}

\newcommand{\mI}{\mathcal{I}}

\newcommand{\fD}{\mathfrak{D}}

\newcommand{\vp}{\mathbf{p}}

\newcommand{\vv}{\mathbf{v}}

\newcommand{\vf}{\mathbf{f}}

\newcommand{\vd}{\mathbf{d}}

\newcommand{\vDelta}{\boldsymbol{\Delta}}

\newcommand{\vg}{\mathbf{g}}

\newcommand{\vD}{\mathbf{D}}
\newcommand{\vzero}{\mathbf{0}}

\newcommand{\vc}{\mathbf{c}}

\newcommand{\vone}{\mathbf{1}}

\newcommand{\vlambda}{\boldsymbol{\lambda}}

\newcommand{\vs}{\mathbf{s}}
\newcommand{\vP}{\mathbf{P}}
\newcommand{\vr}{\mathbf{r}}

\begin{document}

\title{Probabilistic Trajectory Planning for Static and Interaction-aware Dynamic Obstacle Avoidance}

\author{Bask{\i}n \c{S}enba\c{s}lar and
        Gaurav S. Sukhatme
\thanks{Bask{\i}n \c{S}enba\c{s}lar and Gaurav S. Sukhatme are with the Department of Computer Science, University of Southern California, Los Angeles, CA, USA.}
\thanks{Email: \{baskin.senbaslar, gaurav\}@usc.edu}
}

\markboth{}%
{}

\IEEEpubid{}

\maketitle

\begin{abstract}
Collision-free mobile robot navigation is an important problem for many robotics applications, especially in cluttered environments.
In such environments, obstacles can be static or dynamic.
Dynamic obstacles can additionally be interactive, i.e. changing their behavior according to the behavior of other entities.
The perception and prediction modules of robotic systems create probabilistic representations and predictions of such environments.
In this paper, we propose a novel prediction representation for interactive behaviors of dynamic obstacles.
Then, we propose a real-time trajectory planning algorithm that probabilistically avoids collisions against static and interactive dynamic obstacles, and produces dynamically feasible trajectories.
During decision making, our planner simulates the interactive behavior of dynamic obstacles in response to the actions planning robot takes.
We explicitly minimize collision probabilities against static and dynamic obstacles using a multi-objective search formulation.
Then, we formulate a quadratic program to safely fit a smooth trajectory to the search result while attempting to preserve the collision probabilities computed during search.
We evaluate our algorithm extensively in simulations to show its performance under different environments and configurations using 78000 randomly generated cases.
We compare its performance to a state-of-the-art trajectory planning algorithm for static and dynamic obstacle avoidance using 4500 randomly generated cases.
We show that our algorithm achieves up to 3.8x success rate using as low as 0.18x time the baseline uses. 
We implement our algorithm for physical quadrotors, and show its feasibility in the real world.
\end{abstract}


\section {Introduction}

\begin{figure}
    \centering
    \includegraphics[width=\linewidth]{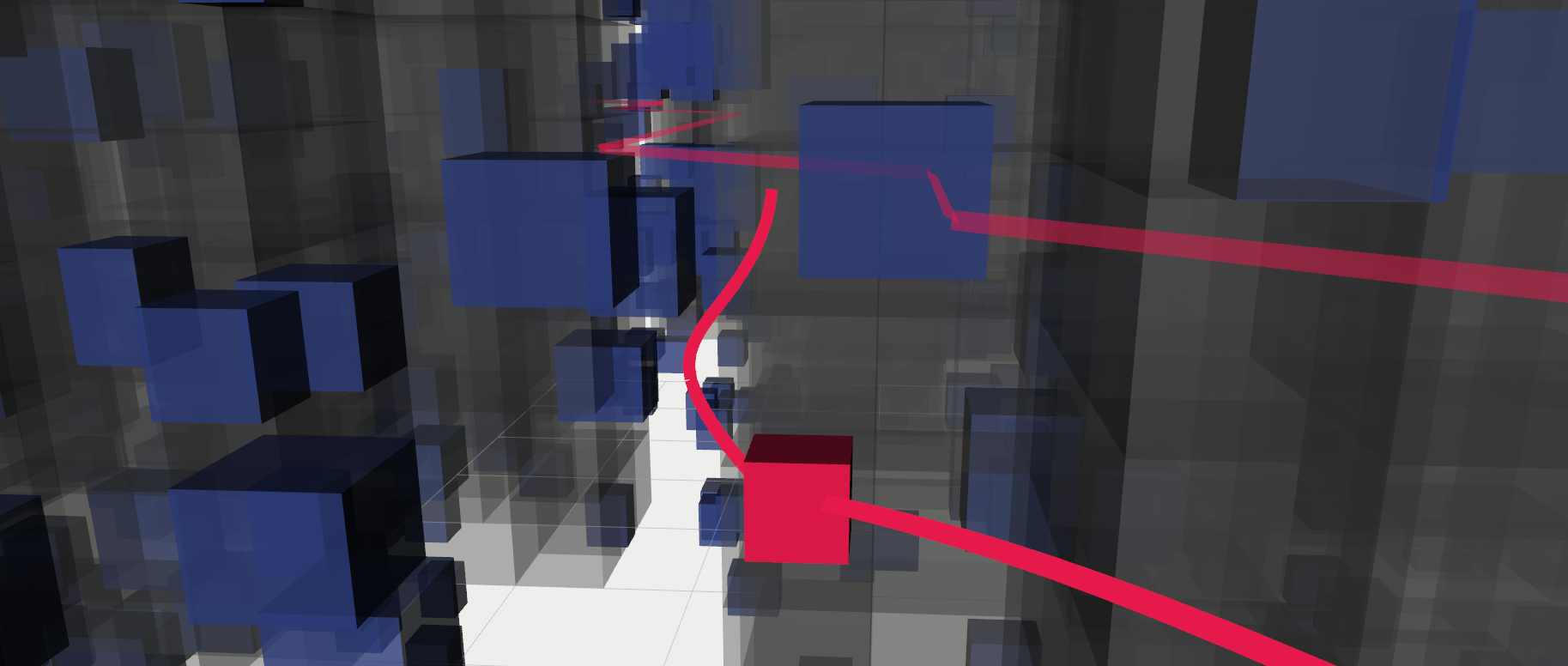}
    \caption{A robot ({\color{red}red}) navigating in a cluttered environment with static ({\color{darkgray}gray}) and dynamic ({\color{blue}blue}) obstacles using our planner.}
    \label{Figure:ClutteredEnvironment}
\end{figure}

Collision-free mobile robot navigation in cluttered environments is an important problem in emerging industries such as autonomous driving~\cite{campbell2010autonomous}, autonomous last-mile delivery~\cite{li2020lastmile}, and human-shared warehouse automation~\cite{inam2018warehouse}.
In such environments, obstacles can be static, i.e. stationary, or dynamic, i.e. moving.
Dynamic obstacles can be interactive, i.e. changing their behavior according to the behavior of other entities.
In this paper, we present a real-time probabilistic trajectory planning algorithm for a mobile robot, which we call the ego robot, that uses its on-board capabilities to navigate in such environments (Fig.~\ref{Figure:ClutteredEnvironment}).

The ego robot uses its on-board sensors to perceive its environment, and classifies obstacles into two sets: static and dynamic.
It produces a probabilistic representation of static obstacles, in which each static obstacle is associated with an existence probability.
It uses a prediction system to predict the possibly interactive behavior models of dynamic obstacles and assigns realization probabilities to each behavior model.

Using these uncertain representations of static and dynamic obstacles, our trajectory planner generates dynamically feasible polynomial trajectories in real-time by primarily minimizing probabilities of collisions against static and dynamic obstacles while minimizing distance, duration, rotations, and energy usage as secondary objectives.
During decision making, we consider interactive behaviors of dynamic obstacles in response to the actions of the ego robot.
The planner is intended to be run in a receding horizon fashion, in which the planned trajectory is executed for a short duration and a new trajectory is planned from scratch.
The planner can be guided with desired trajectories, therefore it can be used in conjunction with offline planners that make longer horizon decision making.

The planner uses a four-stage pipeline similar to~\cite{senbaslar2022rlss}, differing in specific operations at each stage:
\begin{enumerate}
\item Select a goal position on the desired trajectory to plan to as well as the time point the goal position should be (or should have been) reached at,
\item Search a discrete spatio-temporal path to the goal position that minimizes collision probabilities against static and dynamic obstacles as well as total duration, distance, and the number of rotations,
\item Solve a quadratic program to safely fit a smooth trajectory to the discrete plan while attempting to preserve the collision probabilities computed during search,
\item Check for validity of the trajectory for the dynamic limits, and discard the trajectory if not, in which case planning fails and the robot continues using its previously planned trajectory.  
\end{enumerate}

The contributions of our work are as follows:
\begin{itemize}
    \item We define a simple representation for interactive behaviors of dynamic obstacles that can be used within a planner. We propose three model-based prediction algorithms to predict interactive behavior models of dynamic obstacles.
    \item We present a real-time trajectory planning algorithm that probabilistically avoids collisions against static and interactive dynamic obstacles, and produces dynamically feasible polynomial trajectories. 
    \item We evaluate our algorithm extensively in simulations to show its performance under different environments and configurations using $78000$ randomly generated cases. 
    We compare its performance to a state-of-the-art trajectory planning algorithm for static and dynamic obstacle avoidance using $4500$ randomly generated cases, and show that our algorithm achieves up to $3.8$x success rate compared to the baseline using as low as $0.18$x the time the baseline uses.
    We implement our algorithm for physical quadrotors, and show its feasibility in the real world.
\end{itemize}

\section{Related Work}

Collision-free polynomial trajectory generation is studied by several other works.
\cite{richter2013planning} presents a method that uses RRT*~\cite{karaman2010rrtstar} to find a collision-free path against static obstacles kinematically, then solves a quadratic program to smoothen the kinematic path to a continuous piecewise polynomial trajectory that is dynamically feasible.
Collisions are re-checked after optimization.
Additional pieces are added and optimization is re-run until the trajectory is collision-free.
\cite{oleynikova2016quad} proposes a polynomial spline trajectory generation algorithm based on local trajectory optimization in which collision avoidance against static obstacles is integrated into the cost function, hence safety is a soft constraint. 
They utilize Euclidean signed distance transform~\cite{felzenszwalb2012distance} of the environment to compute local collision avoidance gradients.
\cite{chen2016planning} presents a method that finds a shortest path in an environment with static obstacles using standard A* search where octrees~\cite{hornung2013octomap} are used for static obstacle representation.
Then, they compute a safe navigation corridor using the cells of octrees that the path traverses, and compute a smooth piecewise polynomial trajectory contained in the safe navigation corridor.
Similarly,~\cite{liu2017planning} uses jump point search (JPS)~\cite{harabor2011jps} to compute a collision-free discrete path against static obstacles, and constructs safe navigation corridors to optimize a polynomial trajectory within.
\cite{usenko2017real} propose a B-spline trajectory generation algorithm for static obstacle avoidance using only locally build maps using 3D circular buffers.
\cite{gao2018fast} proposes a piecewise polynomial trajectory generation algorithm where pieces are B\'ezier curves~\cite{prautzsch2002bezier} in which they compute an initial collision-free trajectory against static obstacles using fast marching method~\cite{sethian1999fast}, which they consecutively optimize within a safe navigation corridor.
\cite{tordesillas2022faster} proposes a method combining JPS with safe navigation corridor construction and solving an optimization problem to compute a piecewise polynomial, but allowing navigation in unknown environments with static obstacles by computing i) a main trajectory within the known free and unknown spaces, and ii) a backup trajectory within the known free space; using the main trajectory for navigation and falling back to the backup trajectory in case unknown space is detected not free.
\cite{qi2023unstruc} presents a planner avoiding static obstacles as well dynamic obstacles given predicted trajectories of dynamic obstacles, but does not model dynamic obstacle interactivity.

Some planners integrate uncertainty associated with several sources into decision making.
The uncertainty may stem from unmodeled system dynamics, state estimation inaccuracy, perception noise, or prediction inaccuracies.
\cite{tordesillas2020mader} proposes a polynomial trajectory planner that can avoid dynamic obstacles given predicted trajectories of dynamic obstacles along with a maximum possible prediction error.
The dynamic obstacles are assumed to be non-interactive during decision making, in the sense that they do not change their behavior depending on what ego robot does.
Chance constrained RRT (CC-RRT)~\cite{luders2010chance} plans trajectories to avoid dynamic obstacles, conservatively limiting the probability of collisions under Gaussian noise of linear systems and Gaussian noise of dynamic obstacle translation predictions.
\cite{aoude2013probabilistically} presents a trajectory prediction method utilizing Gaussian mixture models to estimate motion models of dynamic obstacles, and using these models within a RRT variant to predict the trajectories of the dynamic obstacles as a set of trajectory particles.
They use these particles within CC-RRT to compute and limit collision probabilities. 
They do not model interactive behavior dynamic obstacles with the ego robot.
~\cite{janson2018monte} presents a Monte Carlo sampling method to compute collision probabilities of trajectories against static obstacles under system uncertainty.
They compute a minimally conservative collision probability bounded trajectory by conducting binary search in robot shape inflation amount and doing planning for each inflation amount.
\cite{zhu2019chance} proposes a chance-constrained MPC formulation for dynamic obstacle avoidance where uncertainty stems from Guassian system model noise, Gaussian state estimation noise and dynamic obstacle model noise.
Dynamic obstacles modelled using constant velocities with Gaussian zero mean acceleration noise.
Interactivity of the dynamic obstacles and the ego robot is not modelled.
\cite{chen2023rast} presents RAST, a risk-aware planner that does not require segmenting obstacles into static and dynamic, but uses a particle based occupancy map~\cite{chen2022continuous} in which each particle is associated with a predicted velocity. 
During planning they compute risk-aware safe navigation corridors by using the predicted number of particles contained in the corridors as a metric for risk.
Dynamic obstacle interactivity is not modelled in RAST.
\cite{nair2022collision} proposes an MPC based collision avoidance method against dynamic obstacles under uncertainty where uncertainty stems from system noise of the ego robot as well as prediction noise for dynamic obstacles.

Predicting future states of dynamical systems is studied extensively.
We list several recent approaches here.
Most of the recent approaches are developed in autonomous vehicles domain.
\cite{wiest2012pred} proposes a trajectory prediction method based on Gaussian mixture models that estimates a Gaussian distribution over future states of a vehicle given its past states.
\cite{lee2017desire} pose future trajectory prediction for multiple dynamic obstacles as an optimization problem on learning a posterior distribution over future dynamic obstacle trajectories given the past trajectories.
They generate multiple predictions for the future using a trained neural network and assign probabilities to each of the predictions.
\cite{kim2022diverse} proposes a multi-modal prediction algorithm for vehicles in which they tackle bias against unlikely future trajectories during training.
\cite{bartoli2018context} proposes a human movement prediction algorithm by utilizing the context information, modelling human-human and human-static obstacle interactions.
\cite{zhou2023dyn} develops a multi-modal pedestrian prediction algorithm utilizing and modelling social interactions between humans as well as human intentions.
The state-of-the-art approaches provide predictions for future trajectories of the dynamic obstacles given past observations, potentially in a multi-modal way, using relatively computationally heavy approaches.
This makes them hard to re-query to model interactivity between the ego robot and the dynamic obstacles during decision making, since each invocation of the predictor incurs a high computational cost.
In this paper, we propose fast to query \emph{policies} as prediction system outputs instead of future trajectories.
The policies model both the intentions of the dynamic obstacles (movement models) and the interaction between dynamic obstacles and the ego robot (interaction models) as vector fields of velocities.
In this sense, they can be considered as artificial potential fields describing the movement of objects in velocity space.

\cite{khatib1986real} introduces artificial potential fields for real-time obstacle avoidance.
In abstract terms, they are functions from states to actions, describing the behavior of robots.
Using artificial potential fields for robot navigation is studied extensively.
In these methods, obstacles are modeled using repulsive fields and navigation goals are modeled using attractive fields.
\cite{rimon1990exact} introduces navigation functions, a special case of artificial potential fields, which can be used for solving the exact robot navigation problem with perfect information where obstacles are spherical.
\cite{ge2002dynamic} applies artificial potential fields to dynamic obstacle avoidance.
When perfect information is not available a priori, local minimums of constructed artificial potential fields may cause robots to deadlock.
\cite{park2003new} propose adding virtual obstacles on local minimas to continue navigation in such situtations.
\cite{fedele2018obstacles} switches between potential fields instead of taking a superposition of them to avoid local minima.
\cite{qidan2006sim} uses simulated annealing to escape local minima.
While we do not use artificial potential fields for decision making of the ego robot, we use vector fields in velocity space to model dynamic obstacle movements, which can be considered artificial potential fields.
Their application to robot navigation is an indication that they can model complex behaviors of dynamic objects.
We use the vector fields to simulate the behavior of dynamic obstacles in response to the actions we plan for the ego robot.

\section{Problem Definition}\label{Section:Problem}

Let $\mR: \mathbb{R}^d \rightarrow P(\mathbb{R}^d)$ be the convex collision shape function of the ego robot, where $\mR(\vp)$ is the subset of $\mathbb{R}^d$ occupied by the robot when placed at position $\vp$.
Here, $d \in \{2,3\}$ is the ambient dimension that ego robot operates in and $P(\mathbb{R}^d)$ is the power set of $\mathbb{R}^d$.
We assume that the ego robot is rigid, and the collision shape function is defined as $\mR(\vp) = \mRx{\vzero} \oplus \{\vp\}$ where $\mRx{\vzero}$ is the shape of the ego robot when placed at position $\vzero$ and $\oplus$ is the Minkowski sum operator.
Note that the shape of the ego robot does not depend on the orientation, which means that either robot's orientation is fixed or its collision shape contains the robot in all orientations.

We assume that the ego robot is differentially flat~\cite{murray1995differential}, i.e., its states and inputs can be expressed in terms of its output trajectory and finite derivatives of it, and the output trajectory is the Euclidean trajectory that the robot follows.
When a system is differentially flat, its dynamics can be accounted for by imposing output trajectory continuity up to required degree of derivatives and imposing constraints on maximum derivative magnitudes throughout the output trajectory.
Many existing systems like quadrotors~\cite{mellinger2011snap} or car-like robots~\cite{murray1993CarLike} are differentially flat.
The ego robot requires output trajectory continuity up to degree $c$, and has maximum derivative magnitude constraints $\gamma_k$ for derivative degrees $k \in \{1, \ldots, K\}$.

The ego robot has a perception and prediction system that detects obstacles in the environment and classifies them into two sets: static obstacles and dynamic obstacles.
Static obstacles are obstacles that do not move.
Dynamic obstacles are obstacles that move with or without interaction with the ego robot.
We do not model interactions between dynamic obstacles and assume that dynamic obstacles only interact with the ego robot.
All obstacles have convex shapes, which are sensed by the perception system.

Perception and prediction systems output static obstacles as a set $\mO$ of convex geometric shapes as well as probabilities for their existence, where $p_{static}(\mQ)$ is the probability that obstacle $\mQ \in \mO$ exists in the environment where $\mQ \subset \mathbb{R}^d$.
Many existing data structures including occupancy grids~\cite{homm2010efficient} and octomaps~\cite{hornung2013octomap} support storing obstacles in this form.

The perception and prediction systems output dynamic obstacles as a set $\mD$, where each dynamic obstacle $\fD \in \mD$ is modeled using i) its current position $\vp^{current}_{\fD}$, ii) its convex collision shape function $\mR_{\fD}: \mathbb{R}^d \rightarrow P(\mathbb{R}^d)$, and iii) a probability distribution over its behavior models $m_{\fD,i}$ where each behavior model is a 2-tuple $m_{\fD,i} = (\mM_{\fD, i}, \mI_{\fD,i})$ such that 
$\mM_{\fD, i}$ is the movement model of the dynamic obstacle, and $\mI_{\fD, i}$ is the interaction model of the dynamic obstacle.
$p_{dynamic}(\fD, i)$ is the probability that dynamic obstacle $\fD$ moves according to behavior model $m_{\fD, i}$ such that $\sum_{i}p_{dynamic}(\fD, i) \leq 1$.

A movement model $\mM: \mathbb{R}^d \rightarrow \mathbb{R}^d$ is a function from dynamic obstacle's position to its desired velocity.
An interaction model $\mI: \mathbb{R}^{4d} \rightarrow \mathbb{R}^d$ is a function describing ego robot-dynamic obstacle interaction of the form $\vv_\fD = \mI(\vp_\fD, \vv'_\fD, \vp, \vv)$.
Its arguments are 4 vectors expressed in the same coordinate frame: position $\vp_\fD$ of the dynamic obstacle, desired velocity $\vv'_\fD$ of the dynamic obstacle (which can be obtained from the movement model), position $\vp$ of the ego robot and velocity $\vv$ of ego robot.
Given these, it outputs the velocity $\vv_\fD$ of the dynamic obstacle.
Notice that interaction models do not model inter dynamic obstacle interactions, i.e., the velocity $\vv_\fD$ dynamic obstacle executes does not depend on the position or velocity of other dynamic obstacles.
This is an accurate assumption in sparse environments where dynamic obstacles are not in close proximity to each other, but an inaccurate assumption in dense environments.
We choose to model interactions this way for computational efficiency, memory efficiency as well as sample efficiency during search: modelling inter dynamic-obstacle interactions would result in a combinatorial explosion of possible dynamic obstacle behaviors since we support multiple hypothesis for each dynamic obstacle.
However, one could also define a single joint interaction model for all dynamic obstacles and do non-probabilistic decision making against dynamic obstacles.
While using only position and velocity to model ego robot-dynamic obstacle interaction is an approximation of the reality, we choose this model because of its simplicity.
This simplicity allows us to use interaction models to update the behavior of dynamic obstacles during discrete search efficiently.
The movement and interaction models are policies describing the intention and the interaction of the dynamic obstacles respectively.\footnote{During planning, we evaluate movement and interaction models sequentially to compute the velocity of the dynamic obstacles.
Therefore, one could also combine movement and interaction models, and have a single behavior model for the purposes of our planner.
We choose to model them separately in order to allow a separate prediction of these models.}

The ego robot optionally has a state estimator that estimates its output derivatives up to derivative degree $c$, where degree $0$ corresponds to position, degree $1$ corresponds to velocity, and so on.
If state estimation accuracy is low, the computed trajectories by the planner can be used to compute the expected derivatives in an open-loop fashion assuming perfect execution.
The $i^{th}$ derivative of ego robot's current output is denoted with $\vs_i$ where $i \in \{0, \ldots, c\}$.
$\mS = \{\vs_0, \ldots, \vs_c\}$ is the full state of the ego robot.
We do not utilize the noise associated with state estimation during planning.

The robot is tasked with following a desired trajectory $\vd(t): [0, T] \rightarrow \mathbb{R}^d$ without colliding with obstacles in the environment, where $T$ is the duration of the trajectory.
We define $\vd(t) = \vd(T)\ \forall t > T$.
The desired trajectory $\vd(t)$ can be computed by a global planner using potentially incomplete prior knowledge about obstacles in the environment.
It does not need to be collision-free with respect to static or dynamic obstacles.
If no such global planner exists, it can be set to a straight line from a start position to a goal position.

\section{Approach}

\begin{figure}
\centering
\begin{tikzpicture}
    \node [block] (goalselection) {Goal Selection\\\vspace*{0.1cm} \fcolorbox{black}{white}{\includegraphics[width=3cm]{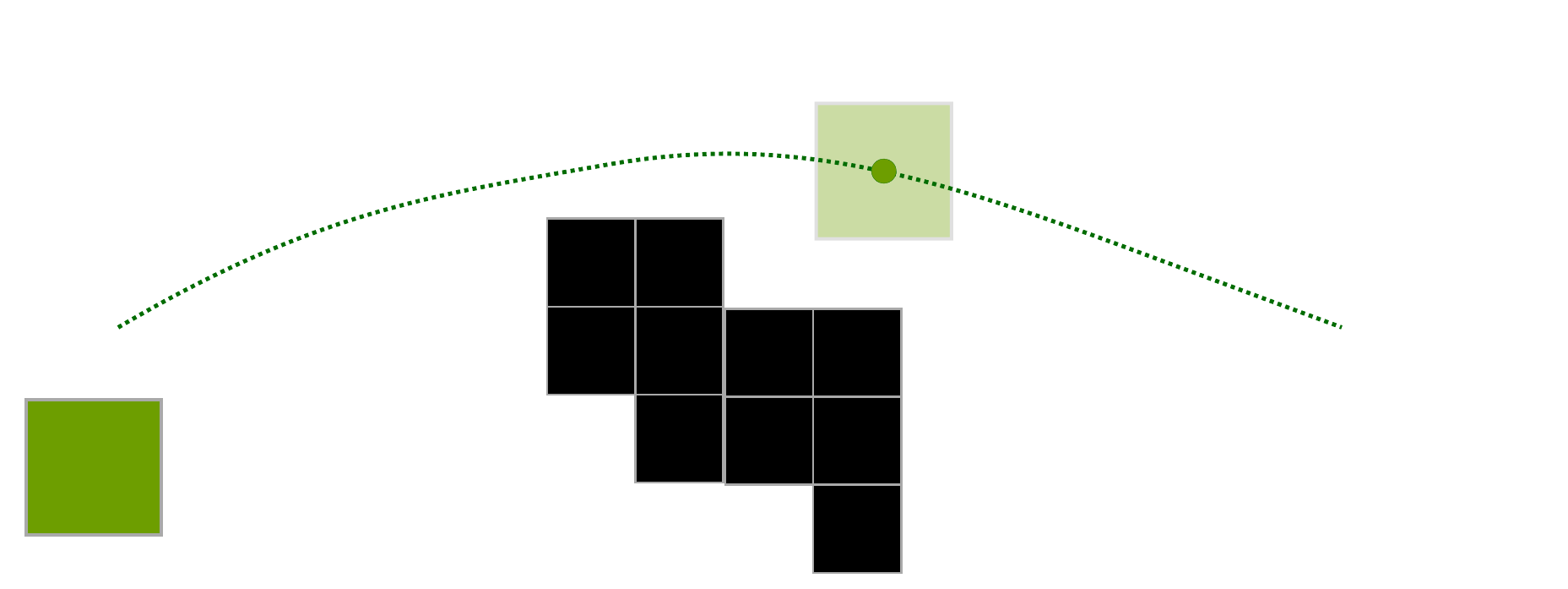}}};
    \node [block, below of = goalselection, node distance=2.5cm] (discretesearch) {Discrete Search\\\vspace*{0.1cm} \fcolorbox{black}{white}{\includegraphics[width=3cm]{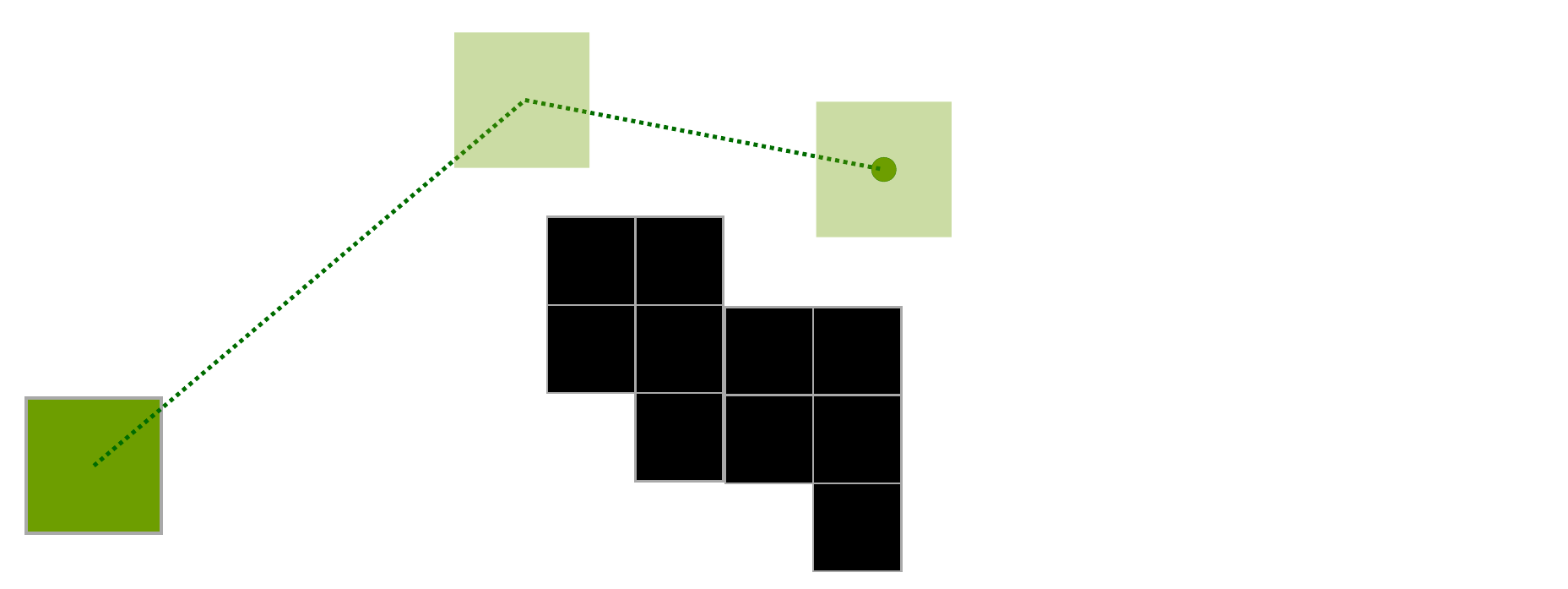}}};
    \node [block, below of = discretesearch, node distance=2.7cm] (trajopt) {Trajectory Optimization\\\vspace*{0.1cm} \fcolorbox{black}{white}{\includegraphics[width=3cm]{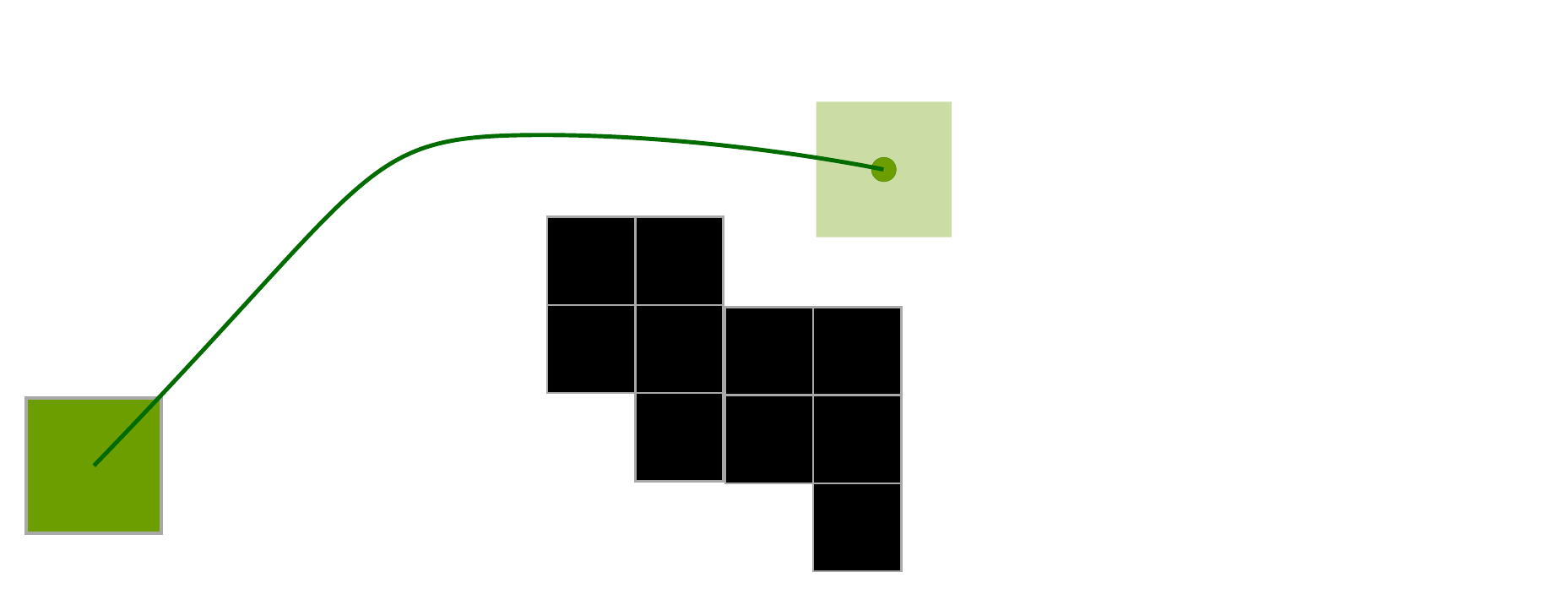}}};
    \node [block, below of = trajopt, node distance=2.5cm] (tempres) {Validity Check};
    \node [empty, below of = tempres, node distance = 1.2cm] (end) {$\vf(t)$};
    \node [empty] at (-2.8, 1.5) (sense) {};
    \node [empty, node distance = 0.5cm] at (-0.4, 1.6) (obstacles) {static \& dynamic obstacles, ego state};
    \node[text width=4cm, color = red] at (-1.3,-8.5) {};
    \path [line] (goalselection) -- node [right, align=left] {goal \& timepoint} (discretesearch);
    \path [line] (discretesearch) -- node [right, align=left] {state sequence} (trajopt);
   \path [line] (trajopt) -- node [right, align=left] {potentially dynamically\\ infeasible trajectory} (tempres);
    \path [line] (tempres) -- (end);
    \path [line, dotted]  (sense) |- (goalselection);
    \path [line, dotted] (sense) |- (discretesearch);
    \path [line, dotted] (sense) |- (trajopt);
\end{tikzpicture}
\caption{Our planning pipeline. Based on the sensed static and dynamic obstacles, and estimated current state of the ego robot, the ego robot computes the trajectory $\vf(t)$ that is dynamically feasible and probabilistically safe.}
\label{Figure:PlanningPipeline}
\end{figure}
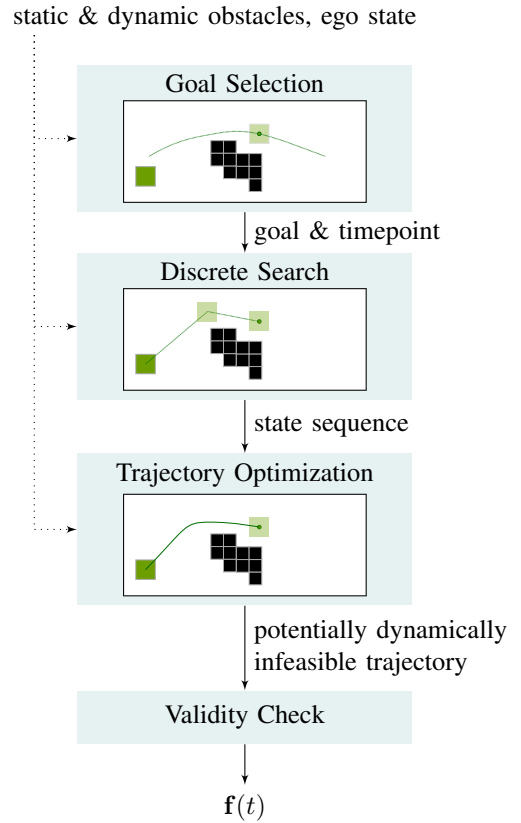

In order to follow the desired trajectory $\vd$ as close as possible while avoiding collisions, we propose a real-time planner that plans for long trajectories which the robot executes for a short duration and re-plans in the next planning iteration.

It is assumed that perception, prediction, and state estimation systems are executed independently from the planner and produce the information described in Section~\ref{Section:Problem}.
To summarize, the inputs from these systems to the planner are:
\begin{itemize}
\item \textbf{Static obstacles}: Set $\mO$ of convex shapes with their existence probabilities such that $p_{static}(\mQ)$ is the probability that obstacle $\mQ \in \mO$ exists.
\item \textbf{Dynamic obstacles}: Set $\mD$ of dynamic obstacles where each dynamic obstacle $\fD\in \mD$ has current position $\vp^{current}_\fD$, collision shape function $\mR_\fD$, and behavior models $m_{\fD, i}$ with corresponding realization probabilities $p_{dynamic}(\fD, i)$.
\item \textbf{Ego robot state}: The state $\mS = \{\vs_0, \ldots, \vs_c\}$ of the ego robot.
\end{itemize}

In each planning iteration, the full planning pipeline (Fig.~\ref{Figure:PlanningPipeline}) is run to compute the next trajectory $\vf(t)$.

There are four stages of our algorithm, which is inspired by~\cite{senbaslar2022rlss}: i) goal selection, which selects a goal position on the desired trajectory to plan to as well as the time it should be (or should have been) reached at, ii) discrete search, which computes a discrete path to the goal position in space-time, minimizing collision probabilities against two classes of obstacles using a multiobjective search method, iii) trajectory optimization, which safely smoothens the discrete path while taking actions to preserve the collision probabilities computed in the discrete search, and iv) validity check, which checks if the dynamic limits of the ego robot is obeyed.
The planner might fail during trajectory optimization, or during validity check, the reasons of which are described in Sections~\ref{Section:TrajectoryOptimization} and~\ref{Section:ValidityCheck}.
If planning fails, robot keeps using its previous plan.
If planning succeeds, it replaces its plan with the new one.

\subsection{Goal Selection}\label{Section:GoalSelection}

Goal selection stage is similar to the one described in~\cite{senbaslar2022rlss}, but we change it for the probabilistic static obstacles. 

In the goal selection stage, we choose a goal position $\vg$ on the desired trajectory $\vd$ to plan to as well as the timepoint $T'$ that the goal position should be (or should have been) reached at.

Goal selection stage has two parameters: Desired planning horizon $\tau$, and minimum static obstacle existence probability $p_{\mO}^{min}$. 
Let $\tilde{T}$ be the current timepoint.

The goal selector finds the timepoint $T'$ that is closest to $\tilde{T} + \tau$ (i.e., the timepoint that is one desired planning horizon away from the current timepoint) when the robot, if placed on the desired trajectory at $T'$, is collision-free against all static obstacles $\mQ \in \mO$ in the environment with existence probability $p_{static}(\mQ) \geq p_{\mO}^{min}$.
Note that goal selection only chooses a single point on the desired trajectory that is collision-free; the actual trajectory the robot follows will be planned by the rest of the algorithm.
Formally, the problem we solve in the goal selection stage is given as follows:
\begin{equation}
    \begin{aligned}
        T' = \arg&\min_{t}  \lvert t-(\tilde{T}+\tau)\rvert \ s.t.\\
        &t\in[0, T]\\
        &\mR(\vd(t)) \cap \mQ = \emptyset\ \forall \mQ\in\mO\ p_{static}(\mQ) \geq p_\mO^{min}. 
    \hspace{-10pt}\end{aligned}
    \label{Equation:GoalSelection}
\end{equation}

We solve~\eqref{Equation:GoalSelection} using linear search on timepoints starting from $\tilde{T} + \tau$ with small increments and decrements.

If there is no safe point on the desired trajectory, i.e. if the robot in a collision state against the objects with more than $p_\mO^{min}$ existence probability when it is placed on any point on the desired trajectory, we return robot's current position $\vs_0$ and current timepoint $\tilde{T}$ as the goal position and timepoint.
This allows us to plan a safe stopping trajectory to the current position of the robot.

\subsection{Discrete Search}\label{Section:DiscreteSearch}

In the discrete search stage, we plan a path to the goal position $\vg$ using cost algebraic $\text{A}^*$ search~\cite{edelkamp2005cost}.
Cost algebraic $\text{A}^*$ search is a generalization of standard $\text{A}^*$ search to a richer set of cost terms, namely cost algebras.
Here, we summarize the formalism of cost algebras from the original paper~\cite{edelkamp2005cost}.
The reader is advised to refer to the original paper for a detailed and complete description of concepts.

\begin{definition}
Let $A$ be a set and $\times: A\times A \rightarrow A$ be a binary operator. A monoid is a tuple $(A, \times, \vone)$  if the identity element $\vone \in A$ exists, $\times$ is associative and $A$ is closed under $\times$.
\end{definition}

\begin{definition}
Let $A$ be a set. A relation $\preceq\ \subseteq A \times A$ is a total order if it is reflexive, anti-symmetric, transitive, and total.
The least operation $\sqcup$ computes the least element of the set according to a total order, i.e. $\sqcup A = c$ such that $c \preceq a\ \forall a\in A$, and the greatest operation $\sqcap$ computes the greatest element of the set according to the total order, i.e. $\sqcap A = c$ such that $a \preceq c\ \forall a \in A$.
\end{definition}

\begin{definition}
A set $A$ is isotone if $a \preceq b$ implies both $a \times c \preceq b \times c$ and $c \times a \preceq c \times b$ for all $a,b,c\in A$.
$a \prec b$ is defined as $a \preceq b \wedge a \neq b$.
A set $A$ is strictly isotone if $a \prec b$ implies both $a \times c \prec b \times c$ and $c\times a \prec c \times b$ for all $a,b,c\in A, c\neq \vzero$ where $\vzero = \sqcap A$.
\end{definition}

\begin{definition}
A cost algebra is a 6-tuple $(A, \sqcup, \times, \preceq, \vzero, \vone)$ such that $(A, \times, \vone)$ is a monoid, $\preceq$ is a total order, $\sqcup$ is the least operation induced by $\preceq$, $\vzero = \sqcap A$, and $\vone = \sqcup A$, i.e. the identity element is the least element.
\end{definition}

Intiutively, $A$ is the set of cost values, $\sqcup$ is the operation used to select the best among the values, $\times$ is the operation to cumulate the cost values, $\preceq$ is the operator to compare the cost values, $\vzero$ is the greatest cost value, and $\vone$ is the least cost value as well as the identity cost value under $\times$.


To support multiple objectives during search, one can define prioritized Cartesian product of cost algebras as follows.

\begin{definition}
    The prioritized Cartesian product of cost algebras $C_1 = (A_1, \sqcup_1, \times_1, \preceq_1, \vzero_1, \vone_1)$ and $C_2 =  (A_2, \sqcup_2, \times_2, \preceq_2, \vzero_2, \vone_2)$, denoted by $C_1 \times_p C_2$ is a tuple $(A_1 \times A_2, \sqcup, \times, \preceq, (\vzero_1, \vzero_2), (\vone_1, \vone_2))$ where $(a_1, a_2) \times (b_1, b_2) = (a_1 \times_1 b_1, a_2 \times_2 b_2)$, $(a_1,a_2) \preceq (b_1,b_2)$ iff $a_1 \prec_1 b_1 \vee (a_1 = b_1 \wedge a_2 \preceq_2 b_2)$, and $\sqcup$ is induced by $\preceq$. 
\end{definition}

\begin{proposition}\label{Proposition:CartesianProductOfCostAlgebras}
If $C_1$ and $C_2$ are cost algebras, and $C_1$ is strictly isotone, then $C_1 \times_p C_2$ is also a cost algebra.
If, in addition, $C_2$ is strictly isotone, $C_1 \times_p C_2$ is also strictly isotone.
\begin{proof}
Given in~\cite{edelkamp2005cost}.
\end{proof}
\end{proposition}

Proposition~\ref{Proposition:CartesianProductOfCostAlgebras} allows one to take Cartesian product of any number of strictly isotone cost algebras and end up with a strictly isotone cost algebra.

Given a cost algebra $C = (A, \sqcup, \times, \preceq, \vzero, \vone)$, cost algebraic A* finds a lowest cost path according to $\sqcup$ between two nodes in a graph where edge costs are elements of set $A$, which are ordered according to $\preceq$ and combined with $\times$, lowest cost value is $\vone$ and the largest cost value is $\vzero$.
Cost algebraic A* uses a heuristic for each node of the graph (similar to standard A*), and cost algebraic A* with re-openings finds cost optimal paths only if the heuristics are admissible.
An admissible heuristic for a node is a cost $h\in A$, which underestimates the cost of the lowest cost path from the node to the goal node according to $\preceq$.

We conduct a multiobjective search, in which each individual cost term is a strictly isotone cost algebra, and optimize over their Cartesian product.
The individual cost terms are defined over two cost algebras, namely $(\mathbb{R}_{\geq 0} \cup \{\infty\}, min, +, \leq, \infty, 0)$, i.e. non-negative real number costs with standard addition and comparison, and $(\mathbb{N} \cup \{\infty\}, min, +, \leq, \infty, 0)$, natural numbers with standard addition and comparison, both of which are strictly isotone.
Therefore, any number of their Cartesian products are also cost algebras by Proposition~\ref{Proposition:CartesianProductOfCostAlgebras}, and hence cost algebraic A* optimizes over them.

The planning horizon of the search is $\tau' = max(\tilde{\tau}, T' - \tilde{T}, \alpha\frac{\normtwo{\vs_0 - \vg}}{\tilde{\gamma_1}})$ where $\tilde{\tau}$ is the minimum search horizon parameter and $\tilde{\gamma_1}$ is the maximum speed for search parameter. 
It is set to the maximum of minimum search horizon, time difference between goal timepoint and current time point, and a multiple of the minimum required time to reach to the goal position $\vg$ from current position $\vs_0$ applying maximum speed $\tilde{\gamma}_1$ where parameter $\alpha \geq 1$.
The planning horizon $\tau'$ is used as a suggestion in the search, and is exceeded if necessary as explained later in this section.

\textbf{States.} The states $x$ in our search formulation have 5 components: i) $x.\vp \in \mathbb{R}^d$ is the position of the state, ii) $x.\vDelta \in \{-1, 0, 1\}^d \setminus \{\vzero\}$ is the direction of the state oriented along ego robot's current velocity $\vs_1$ with a rotation matrix $R_{rot} \in SO(d)$ such that $R_{rot}(1, 0, \ldots, 0)^\top = \frac{\vs_1}{\normtwo{\vs_1}}$, iii) $x.t \in [0, \infty)$ is the time of the state, iv) $x.\mO$ is the set of static obstacles that collide with the path from start state to $x$, and v) $x.\mD$ is the set of dynamic obstacle behavior model--position pairs $(m_{\fD, i}, \vp_{\fD, i})$ such that dynamic obstacle $\fD$ moving according to $m_{\fD, i}$ does not collide the ego robot following the path from start state to $x$, and the dynamic obstacle ends up at position $\vp_{\fD,i}$.

The start state of the search is $x^1$ with components $x^1.\vp = \vs_0$, $x^1.\vDelta = (1, 0, \ldots, 0)^\top$, $x^1.t = 0$, $x^1.\mO$ are set of all obstacles that intersect with $\mR(\vs_0)$, and $x^1.\mD$ contains all mode--position pairs $(m_{\fD,i}, \vp_{\fD}^{current})$ of dynamic obstacles that do not initially collide with ego robot, i.e. $\mR(\vs_0) \cap \mR_{\fD}(\vp_{\fD}^{current}) = \emptyset$.
The goal states are all states $x^g$ that have $x^g.\vp = \vg$.

\textbf{Actions.} There are three action types in our search. Let $x$ be the current state and $x^+$ be the state after applying an action.

\begin{itemize}
\item FORWARD($s$, $t$) moves the current state $x$ to $x^+$ by applying constant speed $s$ along current direction $x.\vDelta$ for time $t$.
The state components change as follows.

\begin{itemize}
    \item $x^+.\vp = x.\vp + R_{rot}\frac{x.\vDelta}{\normtwo{x.\vDelta}}st$
    \item $x^+.\vDelta = x.\vDelta$
    \item $x^+.t = x.t + t$.
    \item We check static obstacles $\mO^+$ colliding with the ego robot with shape $\mR$ linearly travelling from $x.\vp$ to $x^+.\vp$ and set $x^+.\mO = x.\mO \cup \mO^+$.
    \item First, we initialize $x^+.\mD$ to $\emptyset$. Let $(m_{\fD, i}, \vp_{\fD, i}) \in x.\mD$ be a dynamic obstacle behavior model--position pair that does not collide with the state sequence from the start state to $x$.
    Note that robot applies velocity $\vv =  \frac{x^+.\vp - x.\vp}{x^+.t - x.t}$ from state $x$ to $x^+$.
    We get the desired velocity $\vv_{\fD, i}'$ of the dynamic obstacle at time $x.t$ using its movement model: $\vv_{\fD,i}' = \mM_{\fD, i}(\vp_{\fD, i})$.
    The velocity $\vv_{\fD, i}$ of the dynamic obstacle can be computed using the interaction model: $\vv_{\fD, i} = \mI_{\fD, i}(\vp_{\fD,i}, \vv_{\fD,i}', x.\vp, \vv)$.
    We check whether dynamic obstacle shape $\mR_{\fD}$ linearly swept between $\vp_{\fD,i}$ and $\vp_{\fD,i} + \vv_{\fD,i} t$ collides with ego robot shape $\mR$ linearly swept between $x.\vp$ and $x^+.\vp$. (\emph{\color{olive}conservative collision checking})
    If it does not, we add not colliding dynamic obstacle mode by $x^+.\mD = x^+.\mD \cup \{(m_{\fD,i}, \vp_{\fD,i} + \vv_{\fD,i} t)\}$.
    Otherwise, we discard the mode.
\end{itemize}

\item ROTATE($\vDelta'$) changes the current state $x$ to $x^+$ by changing its direction to $\vDelta'$.
It is only available if $x.\vDelta \neq \vDelta'$.
The rotate action is added to penalize turns during discrete search as discussed in the description of costs below.
The state components change as follows.
\begin{itemize}
    \item $x^+.\vp = x.\vp$
    \item $x^+.\vDelta = \vDelta'$
    \item $x^+.t = x.t$
    \item $x^+.\mO = x.\mO$
    \item $x^+.\mD = x.\mD$
\end{itemize}

\item REACHGOAL changes the current state $x$ to $x^+$ by connecting $x.\vp$ to the goal position $\vg$.
The remaining search horizon for robot to reach its goal position is given by $\tau' - x.t$.
Remember that the maximum speed of the ego robot during search is $\tilde{\gamma_1}$.
The robot needs at least $\frac{\normtwo{\vg - x.\vp}}{\tilde{\gamma_1}}$ seconds to reach to the goal position within speed limit.
We set the duration of this action to maximum of these two values: $\max(\tau' - x.t, \frac{\normtwo{\vg - x.\vp}}{\tilde{\gamma_1}})$.
Therefore, the search horizon is merely a suggestion during search, and is exceeded whenever it is not dynamically feasible to reach to the goal position within the search horizon.
The state components change as follows.
\begin{itemize}
    \item $x^+.\vp = \vg$
    \item $x^+.\vDelta = x.\vDelta$
    \item $x^+.t = x.t + \max(\tau' - x.t, \frac{\normtwo{\vg - x.\vp}}{\tilde{\gamma_1}})$
    \item $x^+.\mO$ is computed in the same way as FORWARD.
    \item $x^+.\mD$ is computed in the same way as FORWARD.
\end{itemize}
\end{itemize}
Note that we run interaction models only when ego robot applies a time changing action (FORWARD or REACHGOAL), which is an approximation of reality because dynamic objects can potentially change their velocities between ego robot actions.
We also conduct {\color{olive}\emph{conservative collision checks}} against dynamic obstacles because we do not include the time domain to the collision check and we check whether two swept geometries intersect.
This conservatism allows us to preserve collision probability upper bounds against dynamic obstacles during trajectory optimization as discussed in Section~\ref{Section:TrajectoryOptimization}.

We compute the no collision probability against static obstacles and a lower bound on no collision probability against dynamic obstacles for each state of the search tree in a computationally efficient way as metadata of each state.
We interleave the computation of sets $x.\mO$ and $x.\mD$ with the probability computation.

\subsubsection{Computing No Collision Probability Against Static Obstacles} 

Let $x^{1:n} = x^1, \ldots, x^n$ be a state sequence such that $x^i.t \geq x^{i-1}.t \wedge (x^i.t = x^{i-1}.t \implies x^i.\vp = x^{i-1}.\vp)\ \forall i \in \{2, \ldots n\}$. (Note that our actions result in state sequences in this form.)
Let $\mC_{s}(x^{i:j})$ be the proposition that is true if and only if ego robot following path $(x^i.\vp, x^i.t), \ldots, (x^j.\vp, x^j.t)$ collides with any of the static obstacles in $\mO$, where $j\geq i$.
Event of not colliding with any of the static obstacles while following a prefix of path $x^{1:n}$ is naturally recursive: $\neg\ \mC_{s}(x^{1:i}) = \neg\ \mC_{s}(x^{1:j}) \bigwedge \neg\ \mC_{s}(x^{j:i})\ \forall j\in \{1, \ldots, i\}$.

We compute the probability of not colliding any of the static obstacles for each prefix of the state sequence $x^{1:n}$ during search, and store this probability as a metadata of each state.
Probability of not colliding with static obstacles $\mO$ while traversing the state sequence $x^{1:i}$ is given by $p(\neg\ \mC_{s}(x^{1:i}))$.
\begin{align*}
    p(\neg\ \mC_{s}(x^{1:i})) &= &&p(\neg\ \mC_{s}(x^{1:i-1}) \wedge \neg\ \mC_{s}(x^{i-1:i}))\\
    &= &&p(\neg\ \mC_{s}(x^{1:i-1})) \times\\
    &&&p(\neg\ \mC_{s}(x^{i-1:i})\ |\ \neg\ \mC_{s}(x^{1:i-1}))
\end{align*}

The first term $p(\neg\ \mC_{s}(x^{1:i-1}))$ is the recursive term that can be obtained from the parent state during search.

The second term $p(\neg\ \mC_{s}(x^{i-1:i}) | \neg\ \mC_{s}(x^{1:i-1}))$ is the conditional term that is computed during state expansion.
Let $\mO^{i:j} \subseteq \mO$ be the set of static obstacles that collides with robot traversing $x^{i:j}$ where $i < j$.
Given that the ego robot has not collided while traversing $x^{1:i-1}$ means that no static obstacle that collides with the robot while traversing $x^{1:i-1}$ exists.
Therefore, we compute the conditional probability as the probability that none of the obstacles in $\mO^{i-1:i} \setminus \mO^{1:i-1}$ exists as ones in $\mO^{i-1:i} \cap \mO^{1:i-1}$ do not exist as presumption.
We assume that static obstacles' non-existence events are independent.
Let $E(\mQ)$ be the event that static obstacle $\mQ\in\mO$ exists.
We have

\begin{align*}
p(\neg\ \mC_{s}(x^{i-1:i}) &\ |\  \neg\ \mC_{s}(x^{1:i-1})) \\
&= p\left(\bigwedge_{\mQ \in \mO^{i-1:i} \setminus \mO^{1:i-1}} \neg\ E(\mQ)\right)\\
&= \prod_{\mQ \in \mO^{i-1:i} \setminus \mO^{1:i-1}} p(\neg E(\mQ))\\
&= \prod_{\mQ \in \mO^{i-1:i} \setminus \mO^{1:i-1}} (1-p_{static}(\mQ))
\end{align*}

The key operation for computing  the conditional is computing the set $\mO^{i-1:i} \setminus \mO^{1:i-1}$.
During node expansion, we compute $\mO^{i-1:i}$ by querying the static obstacles for collisions against the region swept by $\mR$ from position $x^{i-1}.\vp$ to $x^i.\vp$.
We obtain $\mO^{1:i-1}$ from the parent state's $x^{i-1}.\mO$.
The no collision probability is computed according to obstacles in $\mO^{i-1:i}\setminus \mO^{1:i-1}$ and $x^{i}.\mO$ is set to $\mO^{1:i} = \mO^{1:i-1} \cup \mO^{i-1:i}$ as described before.

The recursive term $p(\neg\ \mC_{s}(x^{1:1}))$ is initialized for the start state $x^1$ by computing the non-existence probability of obstacles in $x^{1}.\mO$, i.e., $p(\neg\ \mC_{s}(x^{1:1})) = \prod_{\mQ\in x^1.\mO}(1-p_{static}(\mQ))$.

\subsubsection{Computing a No Collision Probability Lower Bound Against Dynamic Obstacles}
No collision probability lower bound computation against dynamic obstacles are done using not colliding dynamic obstacle modes computed during state expansion.

Let $C_d(x^{i:j})$ be the proposition, conditioned on the full path $x^{1:n}$, that is true if and only if the ego robot following the $(x^i.\vp, x^i.t), \ldots, (x^j.\vp, x^j.t)$ portion of $x^{1:n}$ collides with any of the dynamic obstacles in $\mD$ where $i\leq j$.
Similar to static obstacles, event of not colliding with any of the dynamic obstacles while following a prefix of the path $x^{1:n}$ is recursive: $\neg\ C_d(x^{1:i}) = \neg\ C_d(x^{1:j}) \bigwedge \neg\ C_d(x^{j:i})\ \forall j\in \{1, \ldots, i\}$.

The computation of the probability of no collision against dynamic obstacles are formulated similar to the static obstacles.
\begin{align*}
    P(\neg\ \mC_d(x^{1:i})) = &p(\neg\ \mC_d(x^{1:i-1}) \wedge \neg\ \mC_d(x^{i-1: i})) \\
    = &p(\neg\ \mC_d(x^{1:i-1})) \times\\
    &p(\neg\ \mC_d(x^{i-1:i})\ |\ \neg\ \mC_d(x^{1:i-1}))
\end{align*}

The first term $p(\neg\ \mC_d(x^{1:i-1}))$ is the recursive term that can be can be obtained from the parent state during search.

The second term $p(\neg\ \mC_d(x^{i-1:i})\ |\ \neg\ \mC_d(x^{1:i-1}))$ is the conditional term that is computed during state expansion.
Let $C_{d, \fD}(x^{i:j})$ be the proposition, conditioned on the full path $x^{1:n}$, that is true if and only if the ego robot following the $(x^i.\vp, x^i.t), \ldots, (x^j.\vp, x^j.t)$ portion of $x^{1:n}$ collides with dynamic obstacle $\fD$ where $i\leq j$.
We assume independence between no collisions against different dynamic obstacles.
Under this assumption, the conditional term simplifies as follows.
\begin{align*}
    &p(\neg\ \mC_d(x^{i-1:i})\ |\ \neg\ \mC_d(x^{1:i-1})) \\
    &= p\left(\bigwedge_{\fD \in \mD} \neg \ \mC_{d, \fD}(x^{i-1:i})\ |\ \bigwedge_{\fD\in\mD} \neg\ \mC_{d, \fD}(x^{1:i-1})\right)\\
    &= \prod_{\fD \in \mD} p(\neg\ C_{d, \fD}(x^{i-1:i})\ |\ \neg\ C_{d, \fD}(x^{1:i-1}))
\end{align*}

The computation of the term $p(\neg\ C_{d, \fD}(x^{i-1:i})\ |\ \neg\ C_{d, \fD}(x^{1:i-1}))$ for each obstacle $\fD\in\mD$ is done by using $x^{i-1}.\mD$ and $x^{i}.\mD$.
Given that ego robot following states $x^{1:i-1}$ has not collided with dynamic obstacle $\fD$ means that no behavior model of $\fD$ that resulted in a collision while traversing $x^{1:i-1}$ is realized.
We store all non colliding dynamic obstacles behavior models in $x^{i-1}.\mD$ recursively.
Within these, all dynamic obstacle modes that does not collide with the ego robot while traversing from $x^{i-1}$ to $x^{i}$ are stored in $x^{i}.\mD$.
Let $x^i.\fD$ be the set of all behavior model indices of dynamic obstacle $\fD$ that has not collided with $x^{1:i}$.
Probability that ego robot does not collide with dynamic obstacle $\fD$ while traversing from $x^{i-1}$ to $x^{i}$ given that it has not collided with it while travelling from $x^1$ to $x^{i-1}$ is given as
\begin{align*}
    &p(\neg\ C_{d, \fD}(x^{i-1:i})\ |\ \neg\ C_{d, \fD}(x^{1:i-1}))\\
    &=\frac{\sum_{j \in x^i.\fD} p_{dynamic}(\fD, j)}{\sum_{j \in x^{i-1}.\fD}p_{dynamic}(\fD, j)}.
\end{align*}

The computed no collision probabilities are lower bounds because {\color{olive} \emph{collision checks against dynamic obstacles are done conservatively}}, i.e., time domain is not considered during sweep to sweep collision checks.
Conservative collision checks never miss collisions, but may over-report them.

\textbf{Costs.} 
Let $p_{s}(x^i) = 1-p(\neg\ C_s(x^{1:i}))$ be the probability of collision with any of the static obstacle while traversing the state sequence $x^{1:i}$.
Let $p_{d}(x^i) = 1-p(\neg\ C_d(x^{1:i}))$ be an upper bound for the probability of collision with any of the dynamic obstacles while traversing state sequence $x^{1:i}$.
Note that both probabilities can be computed after computing no collision probabilities for each state $x^i$.
We define $P_{s}(t): [0, x^n.t] \rightarrow [0,1]$ of state sequence $x^{1:n}$ as the linear interpolation of $p_{s}$ between states.
\begin{align*}
    P_{s}(t) &= 
    \begin{cases}
        \frac{x^2.t - t}{x^2.t-x^1.t}p_{s}(x^1) \\\ \ \ +\frac{t-x^1.t}{x^2.t-x^1.t}p_{s}(x^2) & x^1.t\leq t <x^2.t\\
        \ldots\ \\
        \frac{x^n.t - t}{x^n.t-x^{n-1}.t}p_{s}(x^{n-1}) \\\ \ \ +\frac{t-x^{n-1}.t}{x^n.t-x^{n-1}.t}p_{s}(x^n) & x^{n-1}.t\leq t \leq x^n.t
        \end{cases}
\end{align*}
We define $P_{d}(t): [0, x^n.t] \rightarrow [0,1]$ of a state sequence $x^{1:n}$ in a similar way using $p_{d}$.

We associate 5 different cost terms to each state $x^i$ in state sequence $x^{1:n}$: i) $\mJ_{static}(x^i) \in [0, \infty)$ is the cumulative static obstacle collision probability defined as $\mJ_{static}(x^i) = \int_0^{x^i.t}P_s(t)dt$, ii) $\mJ_{dynamic}(x^i)\in [0, \infty)$ is the cumulative dynamic obstacle collision probability defined as $\mJ_{dynamic}(x) = \int_0^{x.t}P_d(t)dt$, iii) $\mJ_{distance}(x^i) \in [0, \infty)$ is the distance travelled from start state $x^1$ to state $x^i$, iv) $\mJ_{duration}(x^i) \in [0, \infty)$ is the time elapsed from start state $x^1$ to state $x^i$, and $\mJ_{rotation}(x^i) \in \mathbb{N}$ is the number of rotations from start state $x^1$ to state $x^i$.

We compute the cost terms of the new state $x^+$ after applying actions to current state $x$ as follows.
\begin{itemize}
    \item $\mJ_{static}(x^+) = \mJ_{static}(x) + \int_{x.t}^{x^+.t}P_s(t)dt$
    \item $\mJ_{dynamic}(x^+) = \mJ_{dynamic}(x) + \int_{x.t}^{x^+.t}P_d(t)dt$
    \item $\mJ_{distance}(x^+) = \mJ_{distance}(x) + \normtwo{x^+.\vp - x.\vp}$
    \item $\mJ_{duration}(x^+) = \mJ_{duration}(x) + (x^+.t - x.t)$
    \item $\mJ_{rotation}(x^+) = \mJ_{rotation}(x) + (x.\vDelta \neq x^+.\vDelta\ ?\ 1\ :\ 0)$
\end{itemize}

Lower cost (with respect to standard comparison operator $\leq$) is better in all cost terms.
All cost terms have the minimum cost of $0$ and upper bound cost of $\infty$.
All cost terms are additive using standard addition operator $+$.
$\mJ_{static}, \mJ_{dynamic}, \mJ_{distance}, $ and $\mJ_{duration}$ are cost algebras ($\mathbb{R}_{\geq 0} \cup \{\infty\}$, min, $+$, $\leq$, $\infty$, $0$) and $\mJ_{rotation}$ is cost algebra ($\mathbb{N} \cup \{\infty\}$, min, $+$, $\leq$, $\infty$, $0$) as described in~\cite{edelkamp2005cost}, both of which are strictly isotone.
Therefore, their Cartesian product is a cost algebra as well.
We define the cost algebra we are optimizing over as the Cartesian product of these cost algebras, and define the cost $\mJ(x)$ of each state $x$ as
\begin{align*}
    \mJ(x) = \begin{bmatrix}\mJ_{static}(x)\\ \mJ_{dynamic}(x)\\ \mJ_{distance}(x)\\ \mJ_{duration}(x)\\ \mJ_{rotation}(x)\end{bmatrix}
\end{align*} 
and order cost terms lexicographically as required by Cartesian product cost algebras. 

This ordering induces an ordering between cost terms: we first minimize cumulative static obstacle collision probability, and among the states that minimizes that, we minimize cumulative dynamic obstacle collision probability, and so on.
Hence, safety is the primary; distance, duration, and rotation optimality are the secondary concerns.
Out of safety against static and dynamic obstacles, we prioritize safety against static obstacles as it is a not a harder task than dynamic obstacle avoidance (if one could enforce safety against dynamic obstacles, they could also enforce safety against static obstacles by modeling them as not moving dynamic obstacles).

The heuristic $H(x)$ we use for each state $x$ during search is as follows.
\begin{align*}H(x) = \begin{bmatrix}
H_{static}(x) \\
H_{dynamic}(x)\\
H_{distance}(x)\\
H_{duration}(x)\\
H_{rotation}(x)
\end{bmatrix} = \begin{bmatrix}
P_s(x.t)H_{duration}(x)\\
P_d(x.t)H_{duration}(x)\\
\normtwo{x.\vp - \vg}\\
\max(\tau'-x.t, \frac{H_{distance}(x)}{\tilde{\gamma_1}})\\
0
\end{bmatrix}
\end{align*}

We first compute $H_{distance}(x)$, which we use in computation of $H_{duration}(x)$.
Then, we use $H_{duration}(x)$ during the computation of $H_{static}(x)$ and $H_{dynamic}(x)$.

\begin{proposition}
    All individual heuristics are admissible.
    \begin{proof}
    
    \textbf{Admissibility of $\boldsymbol{H_{distance}}$:} $H_{distance}(x)$ is the Euclidean distance from $x.\vp$ to $g$, and always underestimates the true distance.

    \textbf{Admissibility of $\boldsymbol{H_{duration}}$:} The goal position $\vg$ can be any position in $\mathbb{R}^d$.
    The FORWARD and ROTATE actions can only move robot in a $(3^d-1)$-connected grid. 
    Therefore, probability that robot reaches $\vg$ by only executing FORWARD and ROTATE actions is essentially zero.
    The robot cannot execute any action after REACHGOAL action in an optimal path to a goal state, because the REACHGOAL action already ends in the goal position and any consecutive actions would only increase the total cost.
    Hence, the last action in an optimal path to a goal state should be REACHGOAL.
    
    If the last action while arriving at state $x$ is REACHGOAL, $x.t \geq \tau'$ holds (as REACHGOAL enforces this, see the descriptions of actions). Since $x.\vp = \vg$, $H_{distance}(x) = 0$. Therefore, $H_{duration}(x) = max(\tau' -x.t, \frac{H_{distance}(x)}{\tilde{\gamma_1}}) = 0$, which is trivially admissible, as $0$ is the lowest cost in cost algebra $(\mathbb{R}_{\geq 0} \cup \{\infty\}, min, +, \leq, \infty, 0)$.

    If the last action while arriving at state $x$ is not REACHGOAL, the search should execute REACHGOAL action to reach to the goal position in the future, which enforces that goal position will not be reached before $\tau'$. 
    Also, since the maximum velocity that can be executed during search is $\tilde{\gamma_1}$, robot needs at least $\frac{H_{distance}(x)}{\tilde{\gamma_1}}$ seconds to reach to the goal position.
    Hence, $H_{duration}(x) = max(\tau' - x.t, \frac{H_{distance}(x)}{\tilde{\gamma_1}})$ is admissible.

    \textbf{Admissibility of $\boldsymbol{H_{static}}$ and $\boldsymbol{H_{dynamic}}$:}
    $P_s$ and $P_d$ are not decreasing functions as they are accumulations of linear interpolations of probabilities, which are defined in $[0, 1]$.
    Therefore $P_s(x.t) \leq P_s(t)$ and $P_d(x.t) \leq P_d(t)$ for $t \geq x.t$ in an optimal path to a goal state traversing $x$.
    Since robot needs at least $H_{duration}(x)$ seconds to reach to a goal state, $P_s(x.t)H_{duration}(x)$ underestimates the true static, and $P_d(x.t)H_{duration}(x)$ underestimates the true dynamic obstacle cumulative collision probability from $x$ to a goal state.

    \textbf{Admissibility of $\boldsymbol{H_{rotation}}$:} $H_{rotation} = 0$ is trivially admissible because $0$ is the lowest cost in cost algebra $(\mathbb{N} \cup \{\infty\}, min, +, \leq, \infty, 0)$.
    
    \end{proof}
\end{proposition}

As each individual cost term is admissible, their Cartesian product is also admissible in Cartesian product cost algebra.
Hence, cost algebraic A* with re-openings minimizes over the Cartesian product cost algebra with the given heuristics.


\textbf{Time limited best effort search.} 
Finding the optimal state sequence induced by the costs $\mJ$ to the goal position $\vg$ can take a lot of time.
Because of this, we limit the duration of the search using a maximum search duration parameter $T^{search}$.
When the search gets cut off because of the time limit, we return the lowest cost state sequence to a goal state so far.
During node expansion, A* applies all actions to the state with the lowest cost.
One of those actions is always REACHGOAL.
Therefore, our search formulation connects all expanded states to the goal position using REACHGOAL action.
Hence, when we cut search off, we end up with a lot of candidate plans to the goal position, which are already sorted according to their costs by A*.

We remove the states generated by ROTATE actions from the search result and give the resulting sequence to the trajectory optimization stage.
Note that ROTATE does not change $x.\vp$, $x.t$, $x.\mO$, or $x.\mD$ components of the state and changes only the direction $x.\vDelta$.
The trajectory optimization stage does not use $x.\vDelta$, therefore we remove repeated states from the perspective of trajectory optimization.
Let $x^1,\ldots, x^N$ be the state sequence after this operation, which is fed to trajectory optimization.

\subsection{Trajectory Optimization}\label{Section:TrajectoryOptimization}

In the trajectory optimization stage, we fit a B\'ezier curve $\vf^i(t): [0, T_i] \rightarrow \mathbb{R}^d$ of degree $h_{i}$ where $T_i = x^{i+1}.t - x^i.t$ to each segment from $(x^i.t, x^i.\vp)$ to $(x^{i+1}.t, x^{i+1}.\vp,)$ for all $i \in \{1, \ldots, N-1\}$ to compute a piecewise trajectory $\vf(t):[0, x^N.t]\rightarrow \mathbb{R}^d$ where each piece is the fitted B\'ezier curve, i.e.
\begin{align*}
    \vf(t) = \begin{cases}
        \vf_1(t - x^1.t) & x^1.t = 0 \leq t < x^2.t\\
        \ldots&\ \\
        \vf_{N-1}(t - x^{N-1}.t) & x^{N-1}.t\leq t \leq x^{N}.t
    \end{cases}
\end{align*}
A B\'ezier curve $\vf_i(t)$ of degree $h_{i}$ has $h_{i}+1$ control points $\vP_{i, 0}, \ldots, \vP_{i, h_{i}}$ and is defined as
\begin{align*}
    \vf_i(t) = \sum_{j = 0}^{h_i} \vP_{i, j}{h_i \choose j}\left(\frac{t}{T_i}\right)^j\left(1-\frac{t}{T_i}\right)^{h_i -j}
\end{align*}
The degrees $h_i$ of B\'ezier pieces are parameters.

B\'ezier curves have the convex hull property that they are contained in the convex hull of their control points~\cite{farouki2012bernstein}.
This allows us to easily constrain B\'ezier curves to be inside convex sets by constraining their control points to be in the same convex sets.
Let $\mP = \{\vP_{i, j}\ |\ i\in\{1, \ldots, N-1\}, j\in \{0, \ldots, h_i\} \}$ be the set of the control points of all pieces.

During the trajectory optimization stage, we construct a quadratic program where decision variables are control points $\mP$ of the trajectory.

We do not increase the cumulative collision probabilities $P_{s}$ and $P_{d}$ of the state sequence $x^{1:N}$, by ensuring that robot following $\vf_i(t)$ avoids the same static obstacles and dynamic obstacle behavior models robot travelling from $x^1$ to $x^{i+1}$ avoids.
The constraints generated for this are always feasible.
Note that we do not enforce constraints to $\vf_i(t)$ for static obstacles or dynamic obstacle behavior models that the robot avoids while travelling from $x^i$ to $x^{i+1}$ but does not avoid while travelling from $x^1$ to $x^i$, because doing so would not decrease neither $p_s(x^{i+1})$ nor $p_d(x^{i+1})$ as the robot has already collided with those obstacles.
This has the added benefit of decreasing the number of constraints of the optimization problem.

In order to encourage dynamic obstacles determine their behavior using the interaction model in the same way they determine it in response to $x^{1:N}$, we add cost terms that matches the position and the velocity at the start of each B\'ezier piece $\vf_i(t)$ to the position and the velocity of the robot at $x^{i}$. 

\subsubsection{Constraints}
There are $4$ types of constraints we impose on the trajectory, all of which are linear in control points $\mP$.


\begin{figure}
    \centering
    \includegraphics[width=\linewidth]{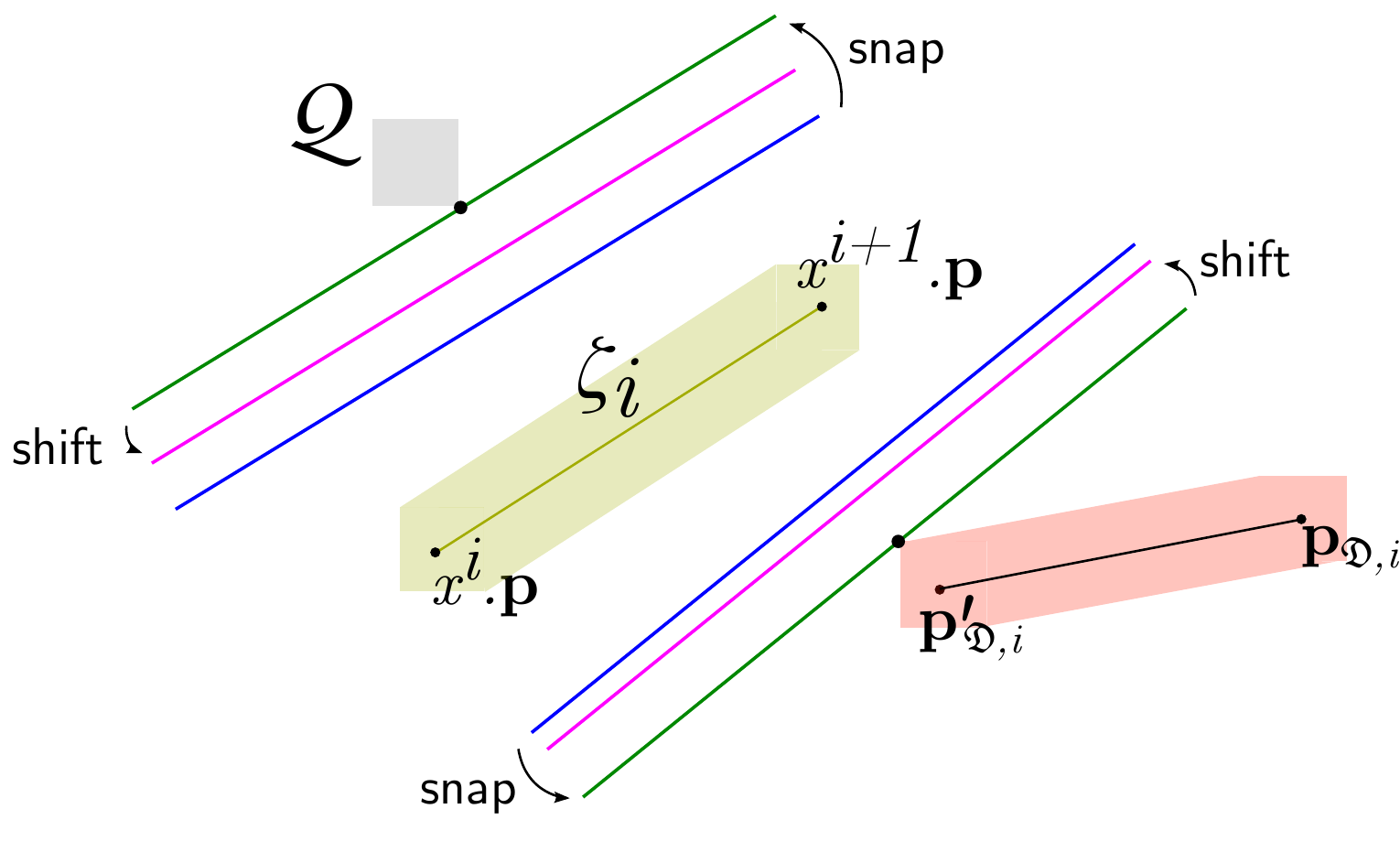}
    \caption{\textbf{Static and dynamic obstacle collision constraints.} Given the {\color{gray}gray} static obstacle $\mQ$ and the {\color{olive}light green} sweep $\zeta_i$ of $\mR$ from $x^i.\vp$ to $x^{i+1}.\vp$, we compute the {\color{blue}blue} support vector machine hyperplane between them. We compute the {\color{ForestGreen}green} separating hyperplane by snapping it to $\mQ$. The robot should stay in the safe side of the {\color{ForestGreen}green} hyperplane. We shift {\color{ForestGreen}green} hyperplane to account for robot's collision shape $\mR$ and compute the {\color{magenta}magenta} hyperplane. The B\'ezier curve $\vf_i(t)$ is constrained by the {\color{magenta}magenta} hyperplane to avoid $\mQ$. To avoid the dynamic obstacle $\fD$ moving from $\vp'_{\fD,i}$ to $\vp_{\fD,i}$, the {\color{blue}} support vector machine hyperplane between the region swept by $\mR_\fD$ and $\zeta_i$ is computed. The same snap and shift operations are conducted to compute the {\color{magenta} magenta} hyperplane constraining $\vf_i(t)$.}
    \label{Figure:SeparatingHyperplanes}
\end{figure}

\textbf{Static obstacle avoidance constraints.}
Let $\mQ \in \mO \setminus x^{i+1}.\mO$ be a static obstacle that robot travelling from $x^1$ to $x^{i+1}$ avoids.
Let $\zeta_i$ be the space swept by the ego robot with shape $\mR$ travelling the straight line from $x^i.\vp$ to $x^{i+1}.\vp$.
Since the shape of the robot is convex and its swept along a straight line segment, $\zeta_i$ is also convex~\cite{senbaslar2022rlss}.
$\mQ$ is also convex by definition.
Since robot avoids $\mQ$, $\mQ \cap \zeta_i = \emptyset$.
Hence, they are linearly separable by the separating hyperplane theorem.
We compute the support vector machine (SVM) hyperplane between $\zeta_i$ and $\mQ$, snap it to $\mQ$, and shift it back to account for ego robot's collision shape $\mR$ as described in~\cite{senbaslar2022rlss} (Fig.~\ref{Figure:SeparatingHyperplanes}).
Let $\mH_{\zeta_i, \mQ}$ be this hyperplane.
We constrain $\vf_i$ with $\mH_{\zeta_i, \mQ}$ for it to avoid $\mQ$, which is a feasible linear constraint as shown in~\cite{senbaslar2022rlss}.

These constraints enforce that robot traversing $\vf_i(t)$ avoids the same obstacles robot traversing from $x^1$ to $x^{i+1}$ avoids, not growing the set $x^{i+1}.\mO$ between $[x^i.t, x^{i+1}.t]\ \forall i\in\{1, \ldots, N-1\}$, and hence $\forall t\in[0, x^N.t]$, preserving the $P_{s}(t)$.

\textbf{Dynamic obstacle avoidance constraints.}
Let $(m_{\fD, i}, \vp_{\fD,i}) \in x^{i+1}.\mD$ be a dynamic obstacle behavior model--position pair that does not collide with robot travelling from $x^{1}$ to $x^{i+1}$.
This means that $m_{\fD, i}$ should be in $x^i$ as well.
Let $\vp'_{\fD, i}$ be the position of the dynamic obstacle behavior model at state $x^i$.
During collision check of state expansion from $x^i$ to $x^{i+1}$, we check whether $\mR_{\fD}$ swept from $\vp'_{\fD, i}$ to $\vp_{\fD,i}$ intersects with $\zeta_i$ and add the mode to $x^{i+1}.\mD$ if they do not.
Since these sweeps are convex sets (because they are sweeps of convex sets along straight line segments), using a similar argument to static obstacle avoidance, they are linearly separable.
We compute the SVM hyperplane between them, snap it to the swept region by dynamic obstacle and shift it to account for the ego robot shape $\mR$.
We constrain $\vf_i$ with this hyperplane, which is a feasible linear constraint as shown in~\cite{senbaslar2022rlss}.

These constraints enforce that robot traversing $\vf_i(t)$ avoids same dynamic obstacle modes robot travelling from $x^1$ to $x^{i+1}$ avoids, not shrinking the set $x^{i+1}.\mD\ \forall i\in\{1,\ldots, N-1\}$, and hence $\forall t\in[0, x^N.t]$, preserving $P_{d}(t)$.

The reason we do {\color{olive}\emph{conservative collision checks}} for dynamic obstacle avoidance during discrete search is to use the separating hyperplane theorem.
Without the conservative collision check, there is no proof of linear separability, and SVM computation might fail.


\textbf{Continuity constraints.}
We enforce continuity up to desired degree $c$ between pieces by
\begin{align*}
    \frac{d^j\vf_i(T_i)}{dt^j} = \frac{d^j\vf_{i+1}(0)}{dt^j}\ \forall i \in \{1,\ldots, N-2\}\ \forall j\in\{0,\ldots,c\}.
\end{align*}

We enforce continuity up to desired degree $c$ between planning iterations by
\begin{align*}
    \frac{d^j\vf_1(0)}{dt^j} = \vs_j\ \forall j\in\{0, \ldots, c\}.
\end{align*}

\textbf{Dynamic limit constraints.}
To enforce dynamic limits during trajectory optimization, we linearly constrain the derivatives of the trajectory in each dimension independently on sampled points by
\begin{align*}
    -\frac{\gamma_k}{\sqrt{d}} \preceq \frac{d^k\vf(j\Delta t)}{dt^k} &\preceq \frac{\gamma_k}{\sqrt{d}}\\
    &\forall j \in \{0, \ldots, \left\lfloor\frac{x^N.t}{\Delta t}\right\rfloor\}\ \forall k \in \{1,\ldots, K\}
\end{align*}
where $\Delta t$ is the sampling interval and $\preceq$ is the component-wise less than operator, such that all elements of a vector is less than (or more than, if the scalar is on the left) the given scalar.

Note that these constraints only enforce dynamic limits at sampled points, and the planned trajectory may be dynamically infeasible between sampled points.
We check for infeasibilities between sampled points during validity check.

\subsubsection{Objective Function}
We use a linear combination of $3$ cost terms as our objective function, all of which are quadratic in control points $\mP$.

\textbf{Energy term.}
We use sum of integrated squared derivative magnitudes as a metric for energy usage similar to~\cite{senbaslar2022rlss, honig2018quadswarms, richter2013planning}, and define the energy usage cost term $\mJ_{energy}(\mP)$ as
\begin{align*}
\mJ_{energy}(\mP) = \sum_{\lambda_j \in \vlambda} \lambda_j \int_0^T\normtwo{\frac{d^j\vf(t)}{dt^j}}^2dt.
\end{align*}

\textbf{Position matching term.}
We add a position matching term $J_{position}(\mP)$ that penalizes distance between piece endpoints and state sequence positions $x^{2}.\vp, \ldots, x^{N}.\vp$.

\begin{align*}
\mJ_{position}(\mP) = \sum_{i\in\{1,\ldots,N-1\}} \theta_i\normtwo{\vf_i(T_i) - x^{i+1}.\vp}^2
\end{align*}
where $\theta_i$s are weight parameters.

\textbf{Velocity matching term.}
We add a velocity matching term $J_{velocity}$ that penalizes divergence from the velocities of the state sequence $x^{1:N}$ at piece start points.

\begin{align*}
    \mJ_{velocity}(\mP) = \sum_{i\in\{1,\ldots,N-1\}} \beta_i\normtwo{\frac{d\vf_i(0)}{dt} - \frac{x^{i+1}.\vp - x^i.\vp}{x^{i+1}.t - x^i}}^2
\end{align*}
where $\beta_i$ are weight parameters.

Position and velocity matching terms encourage matching the positions and velocities of the state sequence $x^{1:N}$ during optimization in order for dynamic obstacles to make similar interaction decisions against the ego robot following trajectory $\vf(t)$ to they do to the ego robot following the state sequence $x^{1:N}$.
One could also add constraints to the optimization problem to exactly match positions and velocities.
Adding position and velocity matching terms as constraints resulted in a high number of optimization infeasibilities in our experiments.
Therefore, we choose to add them to the cost function of the optimization term in the final algorithm.

\subsection{Validity Check}\label{Section:ValidityCheck}

Trajectory optimization stage enforces maximum derivative magnitudes at sampled points, but the resulting trajectory might still be dynamically infeasible between sampled points.
During the validity check stage, we check whether the maximum derivative magnitudes $\gamma_k$ for $k \in \{1, \ldots, K\}$ are satisfied, i.e. whether $\normtwo{\frac{d^k\vf(t)}{dt^k}} \leq \gamma_k\ \forall k\in \{1, \ldots, K\}$, and discard the trajectory otherwise, failing the planning iteration.

If the trajectory $\vf(t)$ passes the validity check, it is dynamically feasible as it obeys maximum derivative magnitudes the robot can execute and is continuous up to degree $c$ as enforced during trajectory optimization, which are the required conditions of dynamic feasibility for differentially flat robots. 

\section{Evaluation Setup}

We evaluate our planner's behavior in perfect execution simulations in 3D, in which the ego robot moves perfectly according to the planned trajectories, and we do not model controller imperfections.
All experiments are conducted in a computer with Intel(R) i7-8700K CPU @3.70GHz, running Ubuntu 20 as the operating system.
The planning pipeline is executed in a single core of the CPU in each planning iteration.
We use CPLEX 12.10 to solve the quadratic programs generated during trajectory optimization stage, including the SVM problems.
The implementation is in C++ for computational efficiency.

\subsection{Mocking Sensing of Static and Dynamic Obstacles}\label{Section:Mocking}
We use octrees from the octomap library~\cite{hornung2013octomap} to represent static obstacles in the environment.
Each axis aligned box with its existence probability stored in the octree is used as a separate obstacle by our planner.
We mock static obstacle sensing imperfections using three operations that we apply to the actual octree representation of the environment:
\begin{itemize}
\item \textbf{increaseUncertainty:} Increases the uncertainty of existing obstacles by moving their existence probability closer to $0.5$.
Let $p \in [0, 1]$ be the existence probability of an obstacle. If $p \leq 0.5$, we sample a probability in $[p, 0.5]$ uniformly and change $p$ to the random sample. Similarly, if $p > 0.5$, we sample a probability in $[0.5, p]$ and change $p$ to the sample.
\item \textbf{leakObstacles($\boldsymbol{p_{leak}}$):} Leaks each obstacle to a neighbouring region with probability $p_{leak}$. Let $p$ be the existence probability of an obstacle. We leak it to a neighbouring region randomly with probability $p_{leak}p$, and increase the existence probability of the neighbouring region by $p_{leak}p$ if the original obstacle is leaked.
\item \textbf{deleteObstacles:} Deletes obstacles randomly according to their non-existence probabilities.
Let $p$ be the existence probability of an obstacle.
We remove it with probability $1-p$.
\end{itemize}

We model dynamic obstacle shapes $\mR_\fD$ as axis aligned boxes.
To simulate imperfect sensing of $\mR_\fD$, we inflate or deflate it in each axis randomly according to a one dimensional $0$ mean Guassian noise with standard deviation $\sigma$.
Note that, we do not explicitly model dynamic obstacle shape sensing uncertainty during planning, yet we still show our algorithm's performance under such uncertainty.
The ego robot is assumed to be noisily sensing the position and velocity of each dynamic obstacle $\fD$ according to a $2d$ dimensional $0$ mean Guassian noise with positive definite covariance $\Sigma \in \mathbb{R}^{2d \times 2d}$.
The first $d$ terms of the noise are applied to the real position and the second $d$ terms of the noise are applied to the real velocity of the dynamic obstacle to compute sensed the position and velocity at each simulation step.
Let $\vp_\fD^{history}$ be the sensed position and $\vv_\fD^{history}$ be the sensed velocity history of the dynamic obstacle $\fD$.
The uncertainty in sensing positions and velocities of dynamic obstacles creates two different uncertainties reflected to the planner: i) prediction inaccuracy increases, ii) current positions $\vp^{current}_\fD$ of dynamic obstacles fed to the planner becomes wrong.
The planner does not explicitly model inaccuracy of individual predictors but it explicitly models uncertainty across models by assigning probabilities to them.
Similarly, the planner does not explicitly model current position uncertainty of dynamic obstacles.
Yet, we still show our algorithm's behavior under such uncertainties.

\subsection{Predicting Behavior Models of Dynamic Obstacles}\label{Section:Prediction}
We introduce $3$ simple online dynamic obstacle behavior model prediction methods to use during evaluation.
More sophisticated behavior prediction methods can be developed and integrated with our planner, which might potentially use domain knowledge about the environment objects exists or handle position and velocity sensing uncertainties explicitly.

\begin{figure}
    \centering
    \subfloat[][Goal attractive model]{\includegraphics[width=0.45\linewidth]{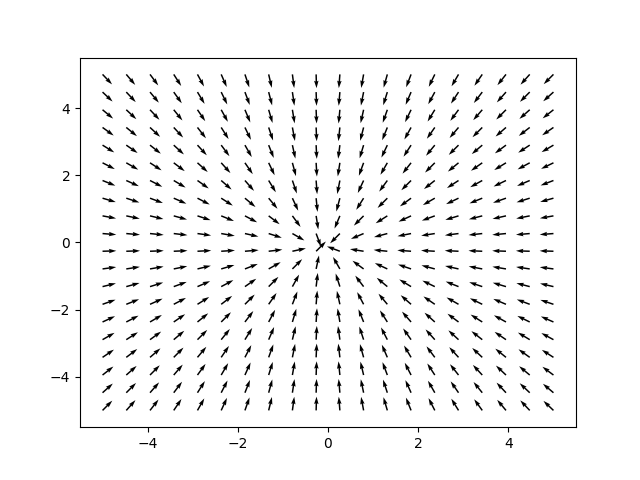}\label{Figure:GoalAttractive}}
    \subfloat[][Constant velocity model]{\includegraphics[width=0.45\linewidth]{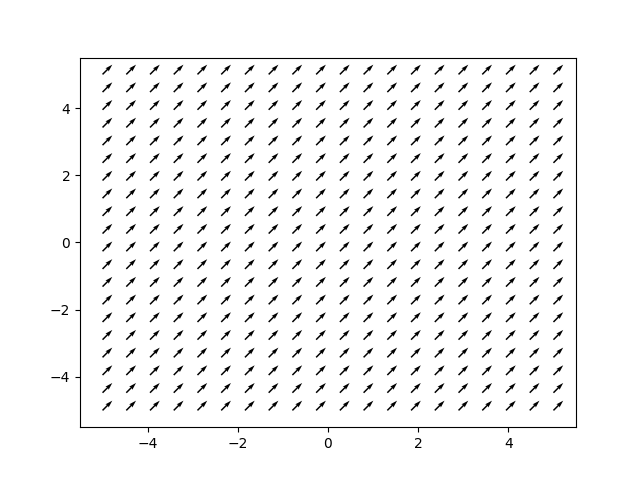}\label{Figure:ConstantVelocity}} \\
    \subfloat[][Rotating model]{\includegraphics[width=0.45\linewidth]{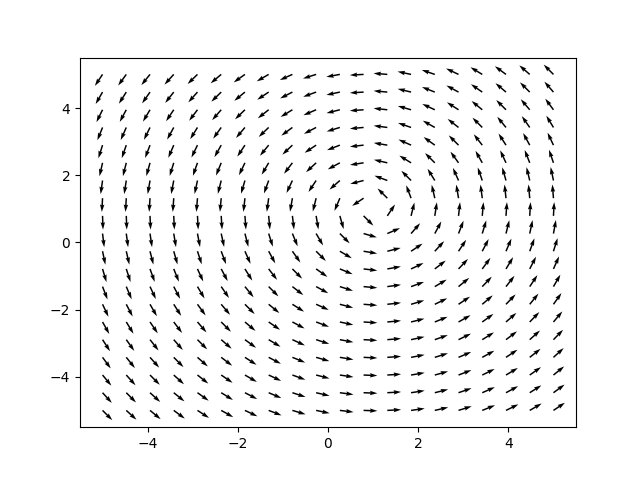}\label{Figure:Rotating}}
    \subfloat[][Repulsive model]{\includegraphics[width=0.45\linewidth]{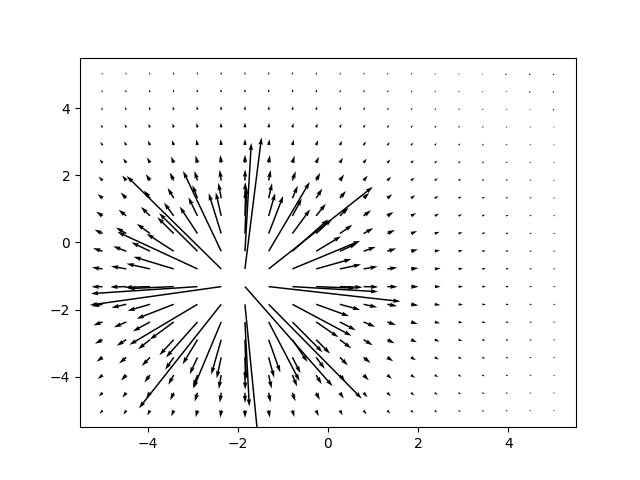}\label{Figure:Repulsive}}
    \caption{The movement and interaction models we define for dynamic obstacles. Each movement and interaction model has associated parameters described in Section~\ref{Section:Prediction}.}
\end{figure}

Let $\vp_\fD$ be the position of dynamic obstacle $\fD$. We define $3$ movement models for dynamic obstacles:
\begin{itemize}
\item \textbf{Goal attractive movement model $\boldsymbol{\mM_g(\vp_\fD | \vg, \hat{s})}$  (Fig.~\ref{Figure:GoalAttractive}):} Attracts the dynamic obstacle to the goal position $\vg$ with desired speed $\hat{s}$.
The desired velocity $\vv_\fD'$ of dynamic obstacle $\fD$ is computed as $\vv_\fD' = \mM_g(\vp_\fD | \vg, \hat{s}) = \frac{\vg - \vp_\fD}{\normtwo{\vg - \vp_\fD}}\hat{s}$.
\item \textbf{Constant velocity movement model $\boldsymbol{\mM_c(\vp_\fD | \vv)}$ (Fig.~\ref{Figure:ConstantVelocity}):} Moves the dynamic obstacle with constant velocity $\vv$.
The desired velocity $\vv_\fD'$ of dynamic obstacle $\fD$ is computed as $\vv_\fD' = \mM_c(\vp_\fD | \vv) = \vv$.
\item \textbf{Rotating movement model $\boldsymbol{\mM_r(\vp_\fD | \vc, \hat{s})}$ (Fig.~\ref{Figure:Rotating}):} Rotates the robot around the rotation center $\vc$ with desired speed $\hat{s}$.
The desired velocity $\vv_\fD'$ of dynamic obstacle $\fD$ is computed as $\vv_\fD' = \mM_r(\vp_\fD | \vc, \hat{s}) = \frac{\vr}{\normtwo{\vr}}\hat{s}$ where $\vr$ is any perpendicular vector to  $\vp_\fD - \vc$.
\end{itemize}
Let $\vp$ be the current position and $\vv$ be the current velocity of the ego robot.
We define $1$ interaction model for dynamic obstacles:
\begin{itemize}
\item \textbf{Repulsive interaction model $\boldsymbol{\mI_r(\vp_\fD, \vv'_\fD, \vp, \vv | f)}$ (Fig.~\ref{Figure:Repulsive}):} Causes dynamic obstacle to repulse away from the ego robot with repulsion strength $f$.
The velocity $\vv_\fD$ of the dynamic obstacle $\vD$ is computed as $\vv_\fD = \mI_r(\vp_\fD, \vv'_\fD, \vp, \vv | f) = \vv'_\fD + \frac{\vp_\fD - \vp}{\normtwo{\vp_\fD - \vp}}\frac{f}{\normtwo{\vp_\fD - \vp}^2}$.
The dynamic obstacle gets repulsed away from the ego robot linearly proportional to repulsion strength $f$, and quadratically inversely proportional to the distance to the ego robot.
\end{itemize}

We use $3$ online prediction methods to predict the behavior models of dynamic obstacles from sensed position and velocity histories of dynamic obstacles as well as the position and velocity histories of the ego robot, one for each combination of movement and interaction models.
Let $\vp^{history}$ be the position and $\vv^{history}$ be the velocity history of the ego robot collected synchronously with $\vp_\fD^{history}$ and $\vv_\fD^{history}$ for each obstacle $\fD$.

\subsubsection{Goal attractive repulsive predictor}
Assuming the dynamic obstacle $\fD$ moves according to goal attractive movement model $\mM_g(\vp_\fD | \vg, \hat{s})$ and repulsive interaction model $\mI_r(\vp_\fD, \vv_\fD', \vp, \vv |f)$, we estimate parameters $\vg$, $\hat{s}$ and $f$ using position and velocity histories of $\fD$ and the ego robot.

During estimation, we solve two consecutive quadratic programs: i) one for goal $\vg$ estimation, ii) one for desired speed $\hat{s}$ and repulsion strength $f$ estimation.
Note that, while joint estimation of $\vg$, $\hat{s}$ and $f$ would be more accurate than estimating them separately, we choose to estimate the parameters in two steps so that the individual problems are QPs, and can be solved fast.
This inaccuracy gets reflected to the probability distribution across behavior models as explained later in this section.

\textbf{Goal estimation.} We estimate the goal position $\vg$ of the dynamic obstacle by computing the point whose average squared distance to rays $\vp_{\fD}^i + t_i\vv_{\fD}^i, t_i \geq 0$, where $\vp_{\fD}^i$ and $\vv_{\fD}^i$ are the elements of $\vp_\fD^{history}$ and $\vv_\fD^{history}$ respectively, is minimal.
Intiutively, $\vp_{\fD}^i + t_i\vv_\fD^i, t_i \geq 0$ is the ray dynamic obstacle would have followed if it did not change its velocity after step $i$.
We compute $\vg$ as the point whose average distance to all these rays is minimum.
The quadratic optimization problem we solve is as follows:
\begin{align*}
    \min_{\vg, t_1, \ldots, t_N} &\frac{1}{N} \sum_{i=1}^{N} \normtwo{\vp_\fD^i + t_i \vv_\fD^i - \vg}^2, \text{s.t.}\\
    &t_i \geq 0\ \forall i\in \{1, \ldots, N\}
\end{align*}
where $N$ is the number of recorded position/velocity pairs.

\textbf{Desired speed and repulsion strength estimation.}
We use the estimated goal position $\vg$, the position history $\vp_{\fD}^{history}$ and the velocity history $\vv_{\fD}^{history}$ of dynamic obstacle, and the position history $\vp^{history}$ and the velocity history $\vv^{history}$ of the ego robot to estimate the desired speed $\hat{s}$ and repulsion strength $f$ of the dynamic obstacle.
Let $\vp^i \in \vp^{history}$ be the position, and $\vv^i \in \vv^{history}$ be the velocity of the ego robot at step $i$.

Assuming the dynamic obstacle moves according to goal attractive repulsive behavior model, its estimated velocity at step $i$ is 
\begin{align*}
    \hat{\vv}_\fD^i &= \mI_r(\vp_{\fD}^i, \mM_g(\vp_{\fD}^i | \vg, \hat{s}), \vp^i, \vv^i)\\
    &= \frac{\vg - \vp_\fD^i}{\normtwo{\vg-\vp_\fD^i}}\hat{s} + \frac{\vp_\fD^i - \vp_i}{\normtwo{\vp_\fD^i - \vp_i}}\frac{f}{\normtwo{\vp_\fD^i - \vp_i}^2}
\end{align*}
We minimize the average squared distance between estimated dynamic obstacle velocities $\hat{\vv}_\fD^i$ and sensed dynamic obstacle velocities $\vv_\fD^i$ to estimate the speed and the repulsion strength:
\begin{align*}
    \min_{s,f} \frac{1}{N}\sum_{i=1}^N \normtwo{\hat{\vv}_\fD^i - \vv_\fD^i}^2
\end{align*}
where $N$ is the number of recorded position/velocity pairs.

\subsubsection{Constant velocity repulsive predictor}
Assuming the dynamic obstacle $\fD$ moves according to the constant velocity movement model $\mM_c(\vp_\fD | \vv)$ and repulsive interaction model $\mI_r(\vp_\fD, \vv_\fD', \vp, \vv |f)$, we estimate the parameters $\vv$ and $f$ using position and velocity histories of $\fD$ and the ego robot.

During estimation, we solve a single quadratic program to estimate both parameters.
Assuming the dynamic obstacle moves according to constant velocity repulsive behavior model, its estimated velocity at step $i$ is 
\begin{align*}
    \hat{\vv}_\fD^i &= \mI_r(\vp_{\fD}^i, \mM_c(\vp_{\fD}^i | \vv), \vp^i, \vv^i)\\
    &= \vv + \frac{\vp_\fD^i - \vp_i}{\normtwo{\vp_\fD^i - \vp_i}}\frac{f}{\normtwo{\vp_\fD^i - \vp_i}^2}
\end{align*}
Similarly to the goal attractive repulsive predictor, we minimize the average squared distance between estimated dynamic obstacle velocities $\hat{\vv}_\fD^i$ and sensed dynamic obstacle velocities $\vv_\fD^i$ to estimate the constant velocity and the repulsion strength:
\begin{align*}
    \min_{\vv,f} \frac{1}{N}\sum_{i=1}^N \normtwo{\hat{\vv}_\fD^i - \vv_\fD^i}^2
\end{align*}
where $N$ is the number of recorded position/velocity pairs.

\subsubsection{Rotating repulsive predictor}
Assuming the dynamic obstacle $\fD$ moves according to the rotating movement model $\mM_r(\vp_\fD | \vc, \hat{s})$ and repulsive interaction model $\mI_r(\vp_\fD, \vv_\fD', \vp, \vv |f)$, we estimate the parameters $\vc$, $\hat{s}$ and $f$ using position and velocity histories of $\fD$ and the ego robot.

We solve two consecutive optimization problems: i) one for rotation center $\vc$ estimation, ii) one for desired speed $\hat{s}$ and repulsion strength $f$ estimation.
We separate estimation to two steps because of a similar reason with the goal attractive repulsive predictor.
The first problem is a linear program and the second one is a quadratic program, both of which allowing fast online solutions.

\textbf{Rotation center estimation.}
We estimate the rotation center $\vc$ of the dynamic obstacle by minimizing averaged dot product between velocity vectors and the vectors pointing from rotation center to the obstacle position.
The reasoning is that if the dynamic obstacle is rotating around a point $\vc$, its velocity vector should always be perpendicular to the vector connecting $\vc$ to obstacle's position.
The linear optimization problem we solve is as follows:
\begin{align*}
    \min_{\vc}\frac{1}{N}\sum_{i=1}^N |\vv_\fD^i \cdot (\vp_\fD^i -\vc)|
\end{align*}
where $N$ is the number of recorded position/velocity pairs. 

\textbf{Desired speed and repulsion strength estimation.}
We use the estimated rotation center $\vc$, as well as position/velocity histories of the dynamic obstacle and the ego robot to estimate the desired speed $\hat{s}$ and repulsion strength $f$ of the dynamic obstacle.

Assuming the dynamic obstacle moves according to rotating repulsive behavior model, its estimated velocity at step $i$ is 
\begin{align*}
    \hat{\vv}_\fD^i &= \mI_r(\vp_{\fD}^i, \mM_r(\vp_{\fD}^i | \vc, \hat{s}), \vp^i, \vv^i)\\
    &= \frac{\vr^i}{\normtwo{\vr^i}}\hat{s} + \frac{\vp_\fD^i - \vp_i}{\normtwo{\vp_\fD^i - \vp_i}}\frac{f}{\normtwo{\vp_\fD^i - \vp_i}^2}
\end{align*}
where $\vr^i$ is a perpendicular vector to $\vp_\fD^i - \vc$.
Similar to the goal attractive repulsive predictor, we minimize the average squared distance between estimated dynamic obstacle velocities $\hat{\vv}_\fD^i$ and sensed dynamic obstacle velocities $\vv_\fD^i$ to estimate the speed and the repulsion strength:
\begin{align*}
    \min_{s,f} \frac{1}{N}\sum_{i=1}^N \normtwo{\hat{\vv}_\fD^i - \vv_\fD^i}^2
\end{align*}
where $N$ is the number of recorded position/velocity pairs.

\subsubsection{Assigning probabilities to predicted behavior models}
The overall prediction system runs these $3$ predictors for each dynamic obstacle $\fD$.
For each predicted behavior model $(\mM_j, \mI_j)$ where $j \in \{1,2,3\}$, we compute the average estimation error $E_j$ as the average $L^2$ norm between the predicted and the actual velocities of the dynamic obstacle.
\begin{align*}
    E_j &= \frac{1}{N}\sum_{i=1}^N \normtwo{\vv_\fD^i - \mI_j(\vp_\fD^i, \mM_j(\vp_\fD^i), \vp^i, \vv^i)}
\end{align*}
We compute the softmax of errors $E_j$ with base $b$ where $0 < b < 1$, and use them as probabilities, i.e. $p_{dynamic}(\fD, j) = \frac{b^{E_j}}{\sum_{i=1}^3 b^{E_i}}$.

\subsection{Evaluation Metrics}
We run each experiment described in Section~\ref{Section:Evaluation} $1000$ times in randomized environments for the statistical significance of results.
In each experiment, the ego robot is tasked with navigating from a start position to a goal position through an environment with static and dynamic obstacles.
There are $8$ metrics that we report for each experiment.

\begin{itemize}
    \item \textbf{Success rate:} Ratio of runs in which robot navigates from its start position to goal position successfully without any collisions.
    \item \textbf{Collision rate:} Ratio of runs in which robot collides with either a static or a dynamic obstacle at least once.
    \item \textbf{Deadlock rate:} Ratio of runs in which robot deadlocks in the environment, i.e. it cannot reach its goal position.
    \item \textbf{Static obstacle collision rate:} Ratio of runs in which robot collides with a static obstacle at least once.
    \item \textbf{Dynamic obstacle collision rate:} Ratio of runs in which robot collides with a dynamic obstacle at least once.
    \item \textbf{Average navigation duration:} Average time it takes for robot to navigate from its start position to its goal position across no-deadlock no-collision runs.
    \item \textbf{Planning fail rate:} Ratio of failing planning iterations over all planning iterations in all runs.
    \item \textbf{Average planning duration:} Average time it takes for one planning iteration over all planning iterations in all runs.
\end{itemize}

\subsection{Fixed Parameters and Run Randomization}

Here, we describe fixed parameters across all experiments, the parameters that are randomized in all experiments, and how they are randomized.

\textbf{Fixed parameters.} 
The minimum static obstacle existence probability $p_\mO^{min}$ of goal selection is set to $0.1$.

The maximum speed $\tilde{\gamma}_1$ of search is set to $\SI{5.0}{\frac{m}{s}}$.
The FORWARD actions of search are FORWARD($\SI{2.0}{\frac{m}{s}}$, $\SI{0.5}{s}$), FORWARD($\SI{3.5}{\frac{m}{s}}$, $\SI{0.5}{s}$), and FORWARD($\SI{4.5}{\frac{m}{s}}$, $\SI{0.5}{s}$).
Minimum search planning horizon $\tilde{\tau}$ is set to $\SI{2.0}{s}$.
Maximum speed duration multiplier $\alpha$ of search is set to $1.5$.

The degrees $h_i$ of B\'ezier curves of trajectory optimization are set to $13$ for all pieces.
The sampling interval $\Delta t$ for dynamic constraints is set to $\SI{0.099}{s}$.
The position matching cost weights $\theta_i$ of trajectory optimization are set to $10, 20, 30$ for the first $3$ pieces, and $40$ for the remaining pieces.
The velocity matching cost weights $\beta_i$ of trajectory optimization are set to $10, 20, 30$ for the first $3$ pieces, and $40$ for the remaining pieces.
The energy term weights of trajectory optimization are set to $\lambda_1 = 2.8$, $\lambda_2 = 4.2$, $\lambda_4 = 0.2$, and $0$ for all other degrees.

In all runs of all experiments, robot navigates in random forest environments, i.e., static obstacles are tree-like objects.
The forest has $\SI{15}{m}$ radius, and trees are $\SI{6}{m}$ high and have a radius of $\SI{0.5}{m}$.
The forest is centered around the origin above $x-y$ plane.
The octree structure we use to represent the static obstacles have a resolution of $\SI{0.5}{m}$.
The density $\rho$ of the forest, i.e., ratio of occupied cells in the octree within the forest, is the only randomization variable about static obstacles.
It is set differently in each experiment as described in Section~\ref{Section:Evaluation}.

\textbf{Run randomization.}
We randomize the following parameters in \emph{all} runs of each experiment in the same way.



\textit{Dynamic obstacle randomization.} 
We randomize the axis aligned box collision shape of each dynamic obstacle by randomizing its size in each dimension uniformly in interval $[\SI{1}{m}, \SI{4}{m}]$. 
The dynamic obstacle's initial position is uniformly sampled in the box $A$ with minimum corner $\begin{bmatrix}\SI{-12}{m} & \SI{-12}{m} & \SI{-2}{m}\end{bmatrix}^\top$ and maximum corner $\begin{bmatrix}\SI{12}{m} & \SI{12}{m} & \SI{6}{m}\end{bmatrix}^\top$.
We sample the movement model of the dynamic obstacle among goal attractive, constant velocity, and rotating models.
If goal attractive movement model is sampled, we sample its goal position $\vg$ uniformly in the same box $A$.
If rotating model is sampled, we sample the rotation center $\vc$ around the origin uniformly in the box with minimum corner $\begin{bmatrix}\SI{-0.5}{m} & \SI{-0.5}{m} & \SI{-0.5}{m}\end{bmatrix}^\top$ and maximum corner $\begin{bmatrix}\SI{0.5}{m} & \SI{0.5}{m} & \SI{0.5}{m}\end{bmatrix}^\top$.
The desired speed $\hat{s}$ is set/randomized differently in each experiment as described in Section~\ref{Section:Evaluation}.
If constant velocity model is sampled, the constant velocity $\vv$ is set by uniformly sampling a unit direction vector and multiplying it with $\hat{s}$.
The interaction model of the dynamic obstacle is always the repulsive model. 
The repulsion strength $f$ is set/randomized differently in each experiment as described in Section~\ref{Section:Evaluation}.
Each dynamic obstacle changes its velocity every decision making period, which is sampled uniformly from interval $[\SI{0.1}{s}, \SI{0.5}{s}]$ for each obstacle.\footnote{Note that, in reality, there is no necessity that the dynamic obstacles move according to the behavior models that we have defined.
The reason we choose to move dynamic obstacles according to these behavior models is so that at least one of our $3$ simple predictors assume the correct model for the dynamic obstacle.
The prediction is still probabilistic, in the sense that we generate $3$ hypothesis for each dynamic obstacle and assign probabilities to each one of them.
One would need to develop new predictors for dynamic obstacles following other behavior models.}

The number of dynamic obstacles $\#_D$ is set/randomized differently in each experiment.


\textit{Robot randomization.}
We randomize the axis aligned box collision shape of each robot by randomizing its size in each dimension uniformly in interval $[\SI{0.2}{m}, \SI{0.3}{m}]$.
We sample the replanning period of the robot uniformly in interval $[\SI{0.2}{s}, \SI{0.4}{s}]$, hence it plans in $\SI{2.5}{Hz}-\SI{5}{Hz}$.
The robot's start position is selected randomly around the forest on a $x-y$ plane aligned circle with radius $\SI{21.5}{m}$ at height $\SI{2.5}{m}$, and the goal position is set to the antipodal point of the start position on the circle.
The robot never collides with a static or a dynamic obstacle initially because static obstacle forest has radius $\SI{15}{m}$, and dynamic obstacles are spawned in a box with corners $\begin{bmatrix}\SI{-12}{m} & \SI{-12}{m} & \SI{-2}{m}\end{bmatrix}^\top$ and $\begin{bmatrix}\SI{12}{m} & \SI{12}{m} & \SI{6}{m}\end{bmatrix}^\top$.

The desired time horizon $\tau$ of goal selection, the time limit $T^{search}$ of search, maximum derivative magnitudes $\gamma_k$ of validity check, required continuity degree $c$ of trajectory optimization, dynamic obstacle position/velocity sensing noise covariance $\Sigma$, and dynamic obstacle shape sensing noise standard deviation $\sigma$ are set/randomized differently in each experiment.

\begin{figure}\label{Figure:SampleEnvironments}
\centering
\subfloat[][$\rho = 0.1, \#_D = 0$]{\includegraphics[width=0.45\linewidth]{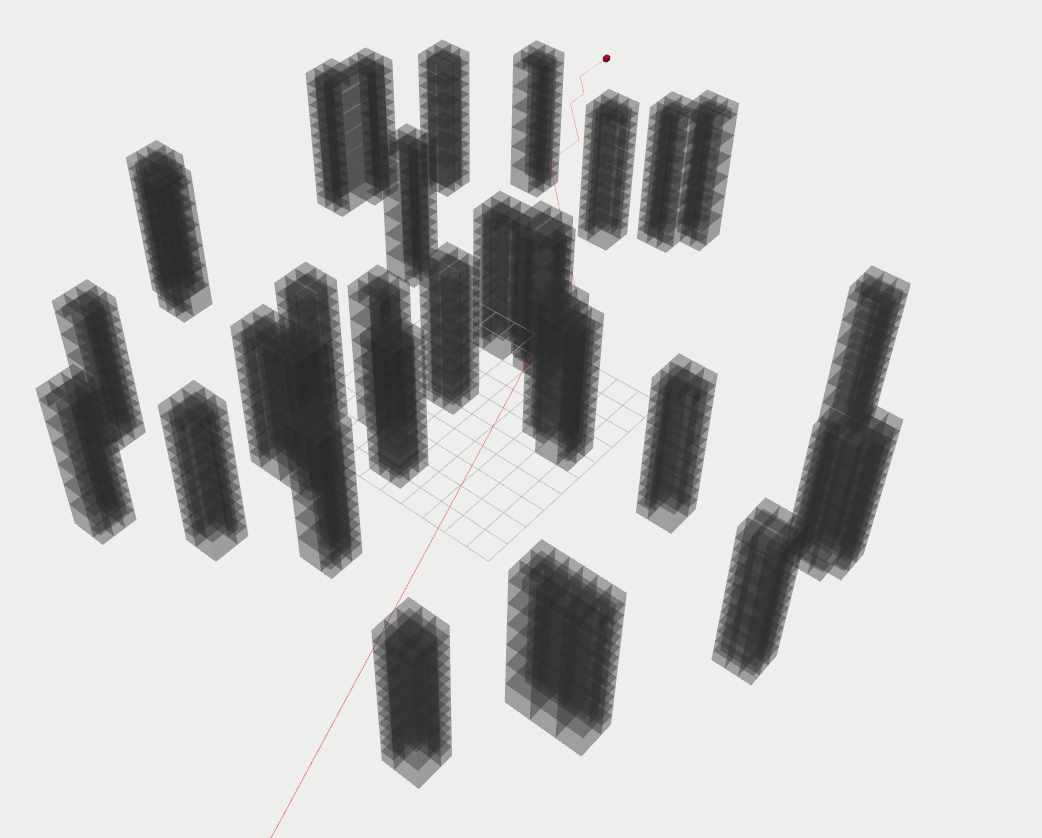}}
\subfloat[][$\rho = 0.2, \#_D = 10$]{\includegraphics[width=0.45\linewidth]{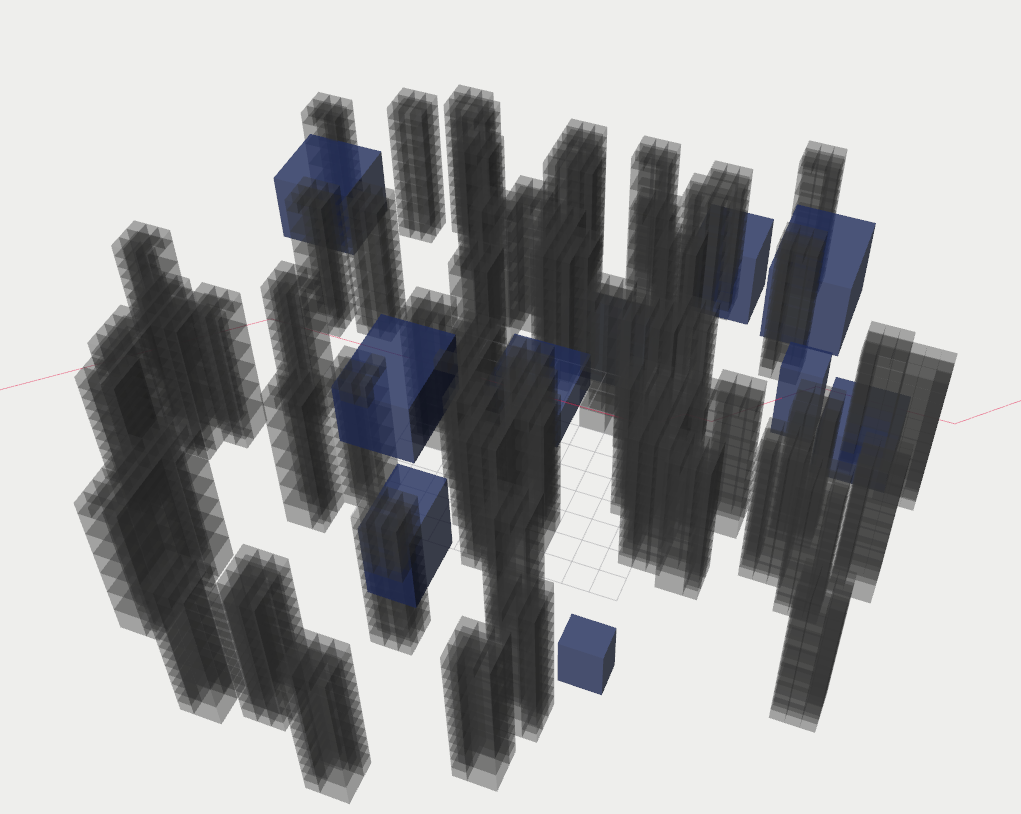}}\\
\subfloat[][$\rho = 0.2, \#_D = 100$]{\includegraphics[width=0.45\linewidth]{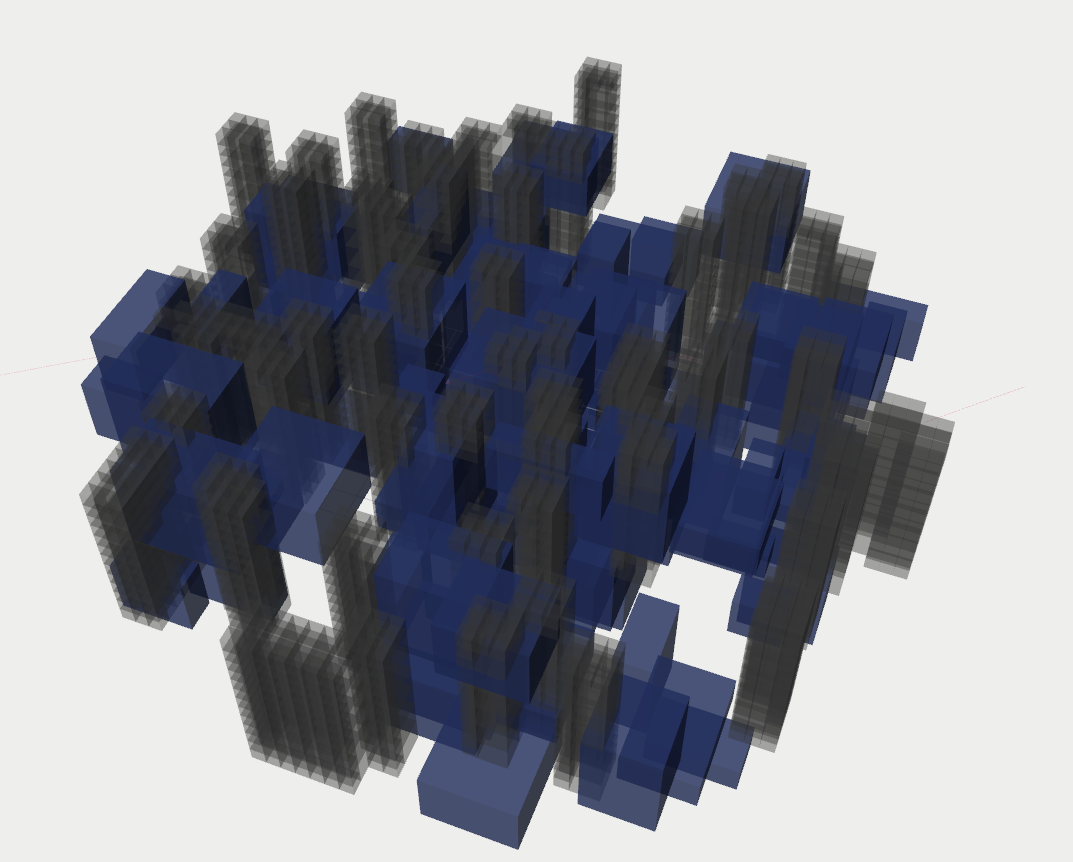}}
\subfloat[][$\rho = 0.0, \#_D = 200$]{\includegraphics[width=0.45\linewidth]{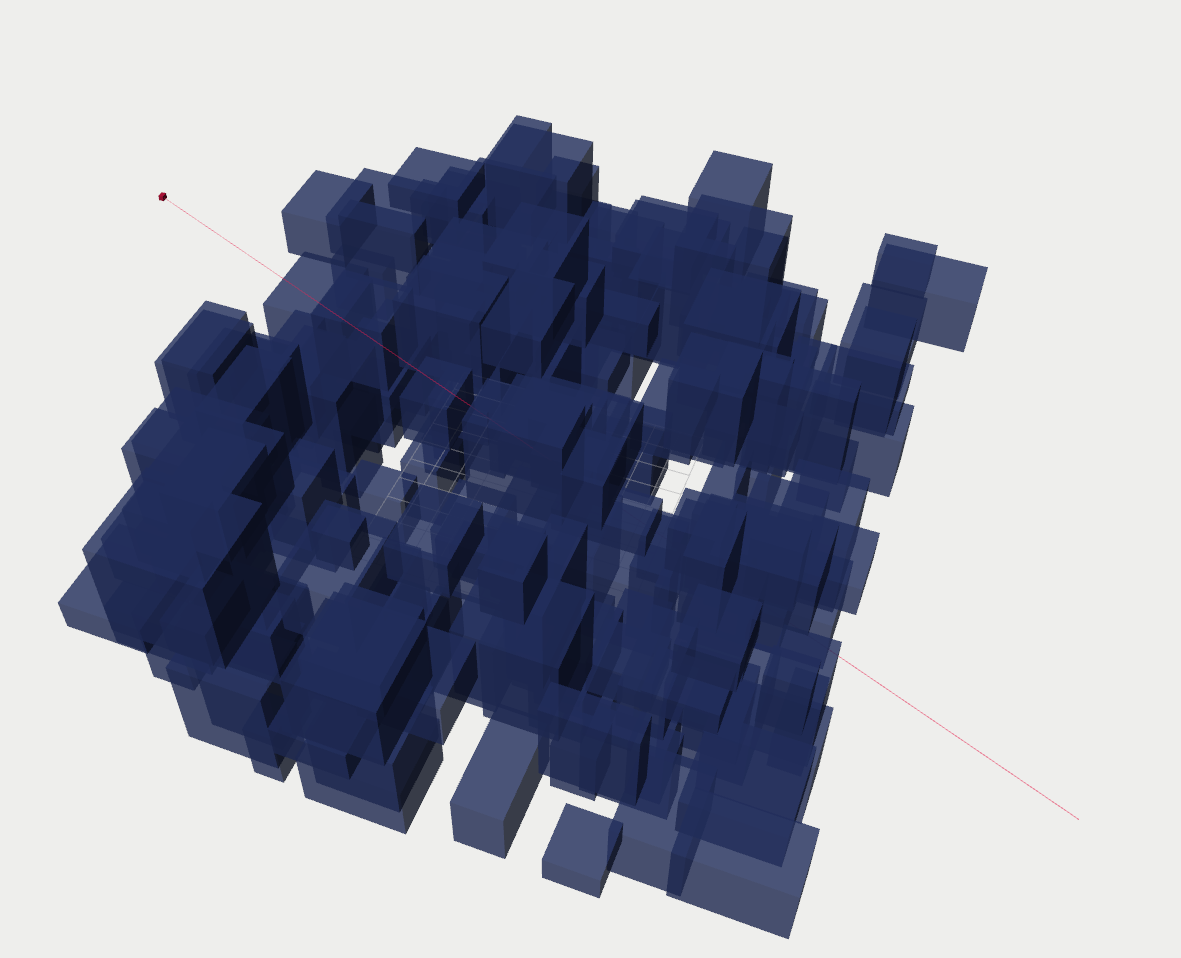}}
\caption{Sample environments generated during run randomization with varying static obstacle density $\rho$ and number of dynamic obstacles $\#_D$. Gray objects are static obstacles representing the forest. Blue obstacles are dynamic obstacles.}
\end{figure}

Sample environments with varying static obstacle densities and number of dynamic obstacles are shown in Fig.~\ref{Figure:SampleEnvironments}.

\section{Evaluation}\label{Section:Evaluation}

We evaluate our algorithm's performance under different configurations and environments in simulations.
We compare our algorithm's performance with MADER~\cite{tordesillas2020mader}.
In addition, we implement our algorithm for physical quadrotors and show its feasibility in real world.

We publish an accompanying video which can be found in \url{https://youtu.be/0XAvpK3qX18}.
The video contains visualization videos from our experiments, as well as visualization of anectodal cases outside of the ones described in this paper.
Also, it contains recordings from our physical experiments.

\subsection{Performance under Different Configurations and Environments}

In this section, we evaluate the performance of our algorithm when it is used in different environments.
In addition, we show its performance under different configurations.

\subsubsection{Desired Trajectory Quality}
To evaluate the effects of desired trajectory quality to navigation performance, we compare two different approaches to compute desired trajectories.
In the first, which we call the \emph{without prior} strategy, we set the desired trajectory of the robot to the straight line segment connecting its start position to its goal position.
In the second, which we call the \emph{with prior} strategy, we set the desired trajectory of the robot to the shortest path between its start position to its goal position avoiding static obstacles.

During these experiments, we set the time limit $T^{search}$ of search to $\SI{75}{ms}$, the required degree of continuity $c = 2$, the maximum velocity magnitudes $\gamma_1 = \SI{10}{\frac{m}{s}}$, the maximum acceleration magnitude of validity check $\gamma_2 = \SI{15}{\frac{m}{s^2}}$.
The desired planning horizon $\tau$ is set to $\SI{2.5}{s}$.
The desired speed $\hat{s}$ of dynamic obstacles is uniformly sampled in interval $[\SI{0.5}{\frac{m}{s}}, \SI{1.0}{\frac{m}{s}}]$, and repulsion strength $f$ uniformly sampled in interval $[0.2, 0.5]$ for each obstacle.
Dynamic obstacle position/velocity sensing noise covariance $\Sigma$, and shape sensing noise standard deviation $\sigma$ are set to $0$, hence the robot senses dynamic obstacles perfectly.
We do not mock sensing of static obstacles, i.e., robot has access to the correct static obstacles in the environment.

The duration of the desired trajectory is set assuming that robot desires to follow it with $\frac{1}{3}$ with its maximum speed $\tilde{\gamma}_1$ of search.

We control the density $\rho$ of the forest, the number $\#_D$ of dynamic obstacles, and whether the desired trajectory is computed using with or without the knowledge of static obstacles a priori, and report the metrics.

The results are summarized in Table~\ref{Table:DesiredTrajectoryQuality}.
With prior strategy results in higher success rates, lower collision rates, deadlock rate, navigation duration, planning failure rate and planning duration in all pairs of cases, beating the without prior strategy in all metrics. 
This suggests the necessity of providing good desired trajectories.

In all following experiments, we use the \emph{with prior} strategy, and provide the desired trajectory by computing the shortest path avoiding only static obstacles and setting its duration assuming that robot desires to follow it with $\frac{1}{3}$ of its maximum speed $\tilde{\gamma}_1$ of search.

\begin{table}[]
    \centering
    \caption{Effects of desired trajectory quality on navigation performance.}
    \label{Table:DesiredTrajectoryQuality}
    \resizebox{\columnwidth}{!}{\begin{tabular}{|c|c|c|c|c|c|c|}
        \hline \textbf{Exp. $\boldsymbol{\#}$} & \textbf{1} & \textbf{2} & \textbf{3} & \textbf{4} & \textbf{5} & \textbf{6}\\
         \hline \textbf{Prior}&  w/o & with & w/o & with & w/o & with\\
         $\boldsymbol{\rho}$ & 0.2 & 0.2 & 0.2 & 0.2 & 0.3& 0.3\\
         $\boldsymbol{\#_D}$& 0 & 0 & 25 & 25& 50& 50\\
        \hline 
         \textbf{succ. rate} & 0.945 & 0.994 & 0.885 & 0.952 & 0.596 & 0.866\\ 
         \textbf{coll. rate} & 0.016 & 0.000 & 0.094 & 0.045 & 0.319 & 0.126\\
        \textbf{deadl. rate}& 0.040 & 0.006 & 0.028 & 0.004& 0.139 & 0.011\\
         \textbf{s. coll. rate}& 0.016 & 0.000 & 0.043 & 0.001 & 0.207 & 0.015\\
         \textbf{d. coll. rate} & 0.000 & 0.000 & 0.062 & 0.044 & 0.217 & 0.118 \\
         \textbf{avg. nav. dur. [s]} & 27.27 & 28.42 & 27.76 & 28.51 & 31.19 & 30.86\\
         \textbf{pl. fail rate}& 0.060 & 0.043  & 0.066 & 0.052 & 0.093 & 0.076\\
         \textbf{avg. pl. dur. [ms]} & 88.02 & 41.14 & 106.27 & 61.42 & 170.56 & 90.41\\
         \hline
    \end{tabular}}
\end{table}

\subsubsection{Enforced Degree of Continuity $c$}

The degree of continuity $c$ we enforce in the trajectory optimization stage determines the smoothness of the resulting trajectory.
The search step enforces only position continuity and degree of continuity greater than that is enforced solely by the trajectory optimizer.

To evaluate the effects of enforced degree of continuity to navigation performance, we compare the navigation metrics when different degrees are enforced.

During these experiments, we do not set any maximum derivative magnitudes to evaluate the effects of only the degree of continuity.
We set $\tau = \SI{2.5}{s}$, $T^{search} = \SI{75}{ms}$, $\Sigma = \vzero$, $\sigma = 0$, $\rho = 0.2$, and $\#_D = 25$.
$\hat{s}$ is sampled in interval $[\SI{0.5}{\frac{m}{s}}, \SI{1.0}{\frac{m}{s}}]$, and repulsion strength $f$ is sampled in interval $[0.2, 0.5]$ uniformly for each obstacle.
We do not mock sensing imperfections of static obstacles.

We control the enforced degree of continuity $c$, and report the metrics.

The results are summarized in Table~\ref{Table:Continuity}.
In general, success rates decrease and deadlock and collision rates increase as the enforced degree of continuity increases.
Interestingly, success rate from degree $0$ to degree $1$ decreases, which we adhere to the non-smooth changes in ego robot's position when $c=0$, causing hard to predict interactions between dynamic obstacles and the ego robot.
Navigation duration is not significantly affected by the enforced degree of continuity.
Planning failure rate increases with the enforced degree of continuity, which is the main cause of collision rate increase.
Average planning duration is not significantly affected by the enforced degree of continuity, because the number of continuity constraints are insignificant compared to safety constraints.

In the remaining experiments, we set $c=2$, i.e., enforce continuity up to acceleration, unless explicitly stated otherwise.

\begin{table}[]
    \centering
    \caption{Effects of enforced continuity degree on navigation performance.}
    \label{Table:Continuity}
    \resizebox{\columnwidth}{!}{\begin{tabular}{|c|c|c|c|c|c|c|}
        \hline \textbf{Exp. $\boldsymbol{\#}$} & \textbf{7} & \textbf{8} & \textbf{9} & \textbf{10} & \textbf{11} & \textbf{12}\\
         \hline $\boldsymbol{c}$ & 0 & 1 & 2 & 3 & 4& 5\\
        \hline 
         \textbf{succ. rate} & 0.980  & 0.986 & 0.971 & 0.930 & 0.893 & 0.830\\ 
         \textbf{coll. rate} & 0.020 & 0.014 & 0.023 & 0.054  & 0.059 & 0.057\\
         \textbf{deadl. rate}& 0.000 & 0.000 & 0.006 & 0.018  & 0.053 & 0.119 \\
         \textbf{s. coll. rate}& 0.001 & 0.000 & 0.000 & 0.000 & 0.000 & 0.003 \\
         \textbf{d. coll. rate} & 0.019 & 0.014 & 0.023  & 0.054 & 0.059 & 0.056 \\
         \textbf{avg. nav. dur. [s]} & 30.87 & 29.23 & 28.57 & 28.50 & 28.31 & 28.29\\
         \textbf{pl. fail rate}& 0.001 & 0.012 & 0.041 & 0.093 & 0.121 & 0.154\\
         \textbf{avg. pl. dur. [ms]} & 49.83 & 51.79 & 53.36  & 55.05 & 53.77 & 54.82\\
         \hline
    \end{tabular}}
\end{table}

\subsubsection{Dynamic Limits}

The dynamic limits $\gamma_k$ of the robot are conservatively enforced during trajectory optimization, and subsequently checked during validity check. 
The planned trajectory is discarded if the dynamic limits are violated.

The discrete search stage limits maximum speed of the resulting discrete plan by enforcing maximum speed $\tilde{\gamma}_1$ (which is set to $\SI{5}{\frac{m}{s}}$ during the experiments).
While this encourages limiting the maximum speed of the resulting trajectory, trajectory optimization stage actually enforces the dynamic limits.

To evaluate the effects of different dynamic limits $\gamma_k$, we compare the navigation metrics when different dynamic limits are enforced.

During these experiments, we set $\tau = \SI{2.5}{s}$, $T^{search} = \SI{75}{ms}$, $\Sigma = \vzero$, $\sigma = 0$, $\rho = 0.2$, and $\#_D = 25$.
The desired speed $\hat{s}$ of dynamic obstacles is sampled in interval $[\SI{0.5}{\frac{m}{s}}, \SI{1.0}{\frac{m}{s}}]$, and repulsion strength $f$ is sampled in interval $[0.2, 0.5]$ uniformly for each obstacle.
We do not mock sensing imperfections of static obstacles.

We control the maximum velocity $\gamma_1$ and maximum acceleration $\gamma_2$, and report the metrics.

The results are given in Table~\ref{Table:DynamicLimits}.
We do not decrease $\gamma_1$ below $\SI{10}{\frac{m}{s}}$, because search has a speed limit of $\tilde{\gamma}_1 = \SI{5}{\frac{m}{s}}$, and setting $\gamma_1 = \SI{10}{\frac{m}{s}}$ enforces $\frac{10}{\sqrt{3}}{\frac{m}{s}} \approx \SI{5.77}{\frac{m}{s}}$ speed limit in all dimensions.
If a lower speed limit is required, maximum speed limit $\tilde{\gamma}_1$ of search should also be decreased.

Unsurprisingly, collision, deadlock and planning failure rates increase and success rate decreases as the dynamic limits get more constraining.
Since the velocity is limited during the discrete search step explicitly, decreasing $\gamma_1$ has a relatively smaller effect on metrics compared to decreasing $\gamma_2$.

In the remaining experiments, we set $\gamma_1 = \SI{10}{\frac{m}{s}}$ and $\gamma_2 = \SI{15}{\frac{m}{s^2}}$, unless explicity stated otherwise.

\begin{table}[]
    \centering
    \caption{Effects of dynamic limits on navigation performance.}
    \label{Table:DynamicLimits}
    \resizebox{\columnwidth}{!}{\begin{tabular}{|c|c|c|c|c|c|c|}
        \hline \textbf{Exp. $\boldsymbol{\#}$} & \textbf{13} & \textbf{14} & \textbf{15} & \textbf{16} & \textbf{17} & \textbf{18}\\
         \hline$\boldsymbol{\gamma_1}[\frac{m}{s}]$ & $\infty$ & $15$ & $10$ & $10$ & $10$ & $10$  \\
         $\boldsymbol{\gamma_2}[\frac{m}{s^s}]$ & $\infty$ & $20$ & $20$ & $15$ & $10$ & $7$   \\
        \hline 
         \textbf{succ. rate} & 0.977 & 0.961 & 0.961 & 0.955 & 0.897 & 0.844\\ 
         \textbf{coll. rate} & 0.022 & 0.035 & 0.037 & 0.037 & 0.101  & 0.148 \\
         \textbf{deadl. rate}& 0.001 & 0.004 &  0.002 & 0.008 & 0.002 & 0.019  \\
         \textbf{s. coll. rate} & 0.002 & 0.002 & 0.000 & 0.000 & 0.007 & 0.013 \\
         \textbf{d. coll. rate} & 0.021 & 0.033 & 0.037 & 0.037 & 0.095 & 0.144 \\
         \textbf{avg. nav. dur. [s]}  &28.55 & 28.57 & 28.51 & 28.56 &   28.49 & 28.47 \\
         \textbf{pl. fail rate} & 0.040 & 0.046 & 0.045 & 0.053  & 0.069  & 0.093 \\
         \textbf{avg. pl. dur. [ms]} & 54.66 & 63.14 & 62.44 & 61.90 &   59.84 & 60.34\\
         \hline
    \end{tabular}}
\end{table}

\subsubsection{Repulsive Dynamic Obstacle Interactivity}

The evaluate the behavior our planner with different levels of repulsive interactivity of dynamic obstacles, we compare navigation metrics when dynamic obstacles use different repulsion strengths.

During these experiments, we set $\tau = \SI{2.5}{s}$, $T^{search} = \SI{75}{ms}$, $\Sigma = \vzero$, $\sigma = 0$, $\rho = 0$, and $\#_D = 50$, i.e., there are no static obstacles and there are $50$ dynamic obstacles in the environment.
The desired speed $\hat{s}$ of dynamic obstacles is sampled in interval $[\SI{0.5}{\frac{m}{s}}, \SI{1.0}{\frac{m}{s}}]$ uniformly for each obstacle.

We control the repulsion strength $f$ and report the metrics.

The results are summarized in Table~\ref{Table:RepulsionStrength}.
In experiment $19$, repulsion strength is set to $-0.5$, causing dynamic obstacles to get attracted to the ego robot, i.e., they move towards the ego robot.
In general, as the repulsive interactivity increases, the collision and deadlock rates decrease, and the success rate increases.
In addition, the average planning duration decreases as the repulsive interactivity increases, because the problem gets easier for the ego robot if dynamic obstacles take some responsibility of collision avoidance even with a simple repulsion rule.

In the remaining experiments, we sample $f$ in interval $[0.2, 0.5]$ uniformly.

\begin{table}[]
    \centering
    \caption{Effects of repulsion strength to navigation performance.}
    \label{Table:RepulsionStrength}
    \resizebox{\columnwidth}{!}{\begin{tabular}{|c|c|c|c|c|c|c|}
        \hline \textbf{Exp. $\boldsymbol{\#}$} & \textbf{19} & \textbf{20} & \textbf{21} & \textbf{22} & \textbf{23} & \textbf{24}\\
         \hline $f$ & $-0.5$ & $0$ & $0.5$ & $1.5$ & $3.0$ & $6.0$  \\
         \hline 
         \textbf{succ. rate} & 0.818 & 0.897 & 0.915 & 0.936 & 0.970 & 0.991\\ 
         \textbf{coll. rate} & 0.182 & 0.102  & 0.084 & 0.064 & 0.030  & 0.009\\
         \textbf{deadl. rate} & 0.001 & 0.004 & 0.002 & 0.000  &  0.000  & 0.000 \\
         {\color{lightgray}\textbf{s. coll. rate}} & {\color{lightgray}0.000} & {\color{lightgray}0.000} & {\color{lightgray}0.000} & {\color{lightgray}0.000} &   {\color{lightgray}0.000} &{\color{lightgray}0.000} \\
         \textbf{d. coll. rate}  & 0.182 & 0.102  & 0.084  & 0.064 &   0.030 & 0.009\\
         \textbf{avg. nav. dur. [s]}   & 26.71 & 26.72 & 26.71 & 26.71 & 26.69  & 26.68\\
         \textbf{pl. fail rate} & 0.029 & 0.030 & 0.026 & 0.022 & 0.019  & 0.014\\
         \textbf{avg. pl. dur. [ms]}  & 49.72 & 50.63 & 50.55  & 46.98  &  43.17 & 37.76\\
         \hline
    \end{tabular}}
\end{table}

\subsubsection{Dynamic Obstacle Desired Speed}
To evaluate the behavior of our planner in environments with dynamic obstacles having different desired speeds, we control the desired speed $\hat{s}$ and report the metrics.

During these experiments, we set $\tau = \SI{2.5}{s}$, $T^{search}=\SI{75}{ms}$, $\Sigma=\vzero$, $\sigma = 0$, $\rho = 0$, $\#_D = 50$.

The results are given in Table~\ref{Table:DesiredSpeed}.
The collision rates increase and the success rate decreases as the desired speed of dynamic obstacles increase.
Deadlock rates are close to $0$ in all cases, because there are no static obstacles in the environment and dynamic obstacles eventually move away from the ego robot.
The average planning duration for the ego robot decreases as the desired speed increases.
We attribute this to the dynamic obstacles with constant velocity.
As the desired speed of the dynamic obstacles increase, constant velocity obstacles leave the environment faster, causing the ego robot to avoid lesser number of obstacles, albeit faster obstacles.
The planning failure rate also increases with the increase in dynamic obstacle desired speeds because the ego robot has to take abrupt actions more frequently to avoid faster obstacles, causing the pipeline to fail because of dynamic limits of the ego robot.

In the remaining experiments we sample desired speed $\hat{s}$ uniformly in interval $[\SI{0.5}{\frac{m}{s}}, \SI{1.0}{\frac{m}{s}}]$.

\begin{table}[]
    \centering
    \caption{Effects of dynamic obstacle desired speed to navigation performance.}
    \label{Table:DesiredSpeed}
    \resizebox{\columnwidth}{!}{\begin{tabular}{|c|c|c|c|c|c|c|}
        \hline \textbf{Exp. $\boldsymbol{\#}$} & \textbf{25} & \textbf{26} & \textbf{27} & \textbf{28} & \textbf{29} & \textbf{30}\\
         \hline $\hat{s} [\frac{m}{s}]$ & $0.5$ & $0.75$ & $1.0$ & $1.5$ & $2.5$ & $5.0$  \\
         \hline 
         \textbf{succ. rate} & 0.937 & 0.914 & 0.900 & 0.864 & 0.717 & 0.272\\ 
         \textbf{coll. rate} & 0.062 & 0.086 & 0.099 & 0.136 & 0.283 & 0.728\\
         \textbf{deadl. rate} & 0.002 & 0.002 & 0.002 & 0.001 & 0.000 & 0.000\\
         {\color{lightgray}\textbf{s. coll. rate}} & {\color{lightgray}0.000} &  {\color{lightgray}0.000}  &  {\color{lightgray}0.000} &   {\color{lightgray}0.000}  & {\color{lightgray}0.000} & {\color{lightgray}0.000}\\
         \textbf{d. coll. rate}  & 0.062 & 0.086 &0.099  &  0.136& 0.283 & 0.728\\
         \textbf{avg. nav. dur. [s]} & 26.68 & 26.74 & 26.72 & 26.67 & 26.68 & 27.36\\
         \textbf{pl. fail rate} & 0.026 & 0.026 & 0.028 & 0.026 & 0.031 & 0.038\\
         \textbf{avg. pl. dur. [ms]} & 48.76 & 51.03 & 48.94 & 36.10 & 30.83 & 25.20\\
         \hline
    \end{tabular}}
\end{table}

\subsubsection{Number of Dynamic Obstacles}
To evaluate the behavior of our planner in environments with different number of dynamic obstacles, we control the number of dynamic obstacles $\#_D$, and report the metrics.

During these experiments, we set $\tau = \SI{2.5}{s}$, $T^{search} = \SI{75}{ms}$, $\Sigma = \vzero$, $\sigma = 0$, and $\rho = 0$. 

The results are given in Table~\ref{Table:NumberOfDynamicObstacles}.
Expectedly, as the number of dynamic obstacles increase, collision, deadlock, and planning failure rates increase, and the success rate decreases.
Average navigation duration is not affected until the number of dynamic obstacles is increased from $100$ to $200$.
Even then, increase in navigation duration is small.

\begin{table}[]
    \centering
    \caption{Effects of number of dynamic obstacles to navigation performance.}
    \label{Table:NumberOfDynamicObstacles}
    \resizebox{\columnwidth}{!}{\begin{tabular}{|c|c|c|c|c|c|c|}
        \hline \textbf{Exp. $\boldsymbol{\#}$} & \textbf{31} & \textbf{32} & \textbf{33} & \textbf{34} & \textbf{35} & \textbf{36}\\
         \hline $\#_D$ & $10$ & $25$ & $50$ & $75$ & $100$ & $200$  \\
         \hline 
         \textbf{succ. rate} & 0.997 & 0.973 & 0.897 & 0.838 & 0.747 & 0.402\\ 
         \textbf{coll. rate} & 0.003 & 0.027 & 0.102 & 0.161 & 0.252 & 0.598\\
         \textbf{deadl. rate} & 0.000 & 0.000 & 0.004 & 0.007 & 0.017 & 0.087\\
         {\color{lightgray}\textbf{s. coll. rate}} & {\color{lightgray}0.000} & {\color{lightgray}0.000} &  {\color{lightgray}0.000} &  {\color{lightgray}0.000}    & {\color{lightgray}0.000} & {\color{lightgray}0.000}\\
         \textbf{d. coll. rate}  & 0.003 & 0.027 &  0.102  &  0.161    & 0.252 & 0.598\\
         \textbf{avg. nav. dur. [s]} & 26.69 &  26.69 &  26.72 &  26.72   & 26.76 & 28.05\\
         \textbf{pl. fail rate} & 0.005 &  0.013  & 0.027  &    0.037 & 0.049  & 0.098\\
         \textbf{avg. pl. dur. [ms]} & 17.88 &  30.72  &  50.59 &  65.90   & 76.68 & 104.70\\
         \hline
    \end{tabular}}
\end{table}

\subsubsection{Static Obstacle Density}
We control the static obstacle density $\rho$ and report the metrics to evaluate the behavior of our planner in environments with different static obstacle densities.

We set $\tau = \SI{2.5}{s}$, $T^{search} = \SI{75}{ms}$, $\Sigma = \vzero$, and $\sigma = 0$. We set number of dynamic obstacles $\#_D = 0$ to evaluate the affects of static obstacles only.
Note that, during these experiments, desired trajectory is set using the \emph{with prior} strategy, meaning that the desired trajectory avoids all static obstacles.
We do not mock sensing imperfections of static obstacles.

The results are given in Table~\ref{Table:StaticObstacleDensity}.
The effect of $\rho$ to success, collision, and deadlock rates is small.
This stems from the fact that desired trajectory is already good, and our planner has to track it as close as possible.
Average navigation duration increases with $\rho$ because the environment gets more complicated, causing the robot to traverse longer paths.
From $\rho=0.1$ to $\rho=0.4$, planning failure rate and planning duration increase, while from $\rho = 0.4$ to $\rho = 0.6$, they decrease.
The reason is the fact that as the environment density increases, it is less likely that the shortest path from the start position to the goal position goes through the forest, as there is no path through the forest..
Instead, desired trajectory avoids the forest all together.
This causes planning to become easier.

\begin{table}[]
    \centering
    \caption{Effects of static obstacle density to navigation performance.}
    \label{Table:StaticObstacleDensity}
    \resizebox{\columnwidth}{!}{\begin{tabular}{|c|c|c|c|c|c|c|}
        \hline \textbf{Exp. $\boldsymbol{\#}$} & \textbf{37} & \textbf{38} & \textbf{39} & \textbf{40} & \textbf{41} & \textbf{42}\\
         \hline $\rho$ & $0.1$ & $0.2$ & $0.3$ & $0.4$ & $0.5$ & $0.6$  \\
         \hline 
         \textbf{succ. rate} & 0.998 & 0.998 & 0.999 & 0.985 & 0.984 & 0.991\\ 
         \textbf{coll. rate} & 0.000  & 0.000 & 0.001 & 0.004 & 0.005 & 0.000\\
         \textbf{deadl. rate} & 0.002 & 0.002 & 0.000 & 0.012 & 0.013 & 0.009\\
         \textbf{s. coll. rate} & 0.000 & 0.000 & 0.001 & 0.004 & 0.005 & 0.000\\
         {\color{lightgray}\textbf{d. coll. rate}} & {\color{lightgray}0.000} & {\color{lightgray}0.000} & {\color{lightgray}0.000}  & {\color{lightgray}0.000} & {\color{lightgray}0.000} & {\color{lightgray}0.000}\\
         \textbf{avg. nav. dur. [s]} & 27.52 & 28.46 & 30.28 & 33.42 & 35.13 & 35.35\\
         \textbf{pl. fail rate} & 0.012 & 0.039 & 0.068 & 0.063 & 0.026 & 0.013\\
         \textbf{avg. pl. dur. [ms]} & 34.33 & 41.73 & 46.73 & 50.12 & 42.74 & 38.98\\
         \hline
    \end{tabular}}
\end{table}

\subsubsection{Time Limit of Search} 

\begin{table}[]
    \centering
    \caption{Effects of the time limit of search with dynamic limits to navigation performance.}
    \label{Table:TimeLimitOfSearch}
    \resizebox{\columnwidth}{!}{\begin{tabular}{|c|c|c|c|c|c|c|}
        \hline \textbf{Exp. $\boldsymbol{\#}$} & \textbf{43} & \textbf{44} & \textbf{45} & \textbf{46} & \textbf{47} & \textbf{48}\\
         \hline $T^{search} [ms]$ & $75$ & $150$ & $300$ & $600$ & $1000$ & $2000$  \\
         \hline 
         \textbf{succ. rate} & 0.655 & 0.731 & 0.708 & 0.730 & 0.682 & 0.700 \\ 
         \textbf{coll. rate} & 0.343 & 0.268 & 0.289 & 0.269 & 0.315 & 0.296\\
         \textbf{deadl. rate} & 0.013 & 0.013 & 0.014 & 0.018 & 0.026 & 0.022\\
         \textbf{s. coll. rate} & 0.031 & 0.016 & 0.005 & 0.003 & 0.001 & 0.000\\
         \textbf{d. coll. rate} & 0.339 & 0.268 & 0.288 & 0.269 & 0.315 & 0.296\\
         \textbf{avg. nav. dur. [s]} & 29.15 & 29.11 & 29.05 & 28.81 & 420.97 & 28.83\\
         \textbf{pl. fail rate} & 0.078 & 0.082 & 0.089 & 0.107 & 0.119 & 0.130\\
         \textbf{avg. pl. dur. [ms]} & 108.51 & 145.55 & 205.40 & 300.02 & 420.97 & 688.66\\
         \hline
    \end{tabular}}
\end{table}

The time limit $T^{search}$ of the discrete search determines the number of states discovered and expanded during A* search, directly affecting the quality of the resulting plan.
To evaluate the affects of $T^{search}$, we run our algorithm with different values of it and report the metrics.
During evaluations, we set $T^{search}$ to values from $\SI{75}{ms}$ to $\SI{2000}{ms}$.
Note that, for high values of $T^{search}$, the planning is not real-time anymore.
However, we still evaluate the hypothetical performance of our algorithm by freezing the simulation until the planning is done.

During these experiments we set $\tau = \SI{2.5}{s}$, $\Sigma = \vzero$, $\sigma = 0$, $\rho = 0.2$, and $\#_D = 100$.
We do not mock sensing imperfections of static obstacles.

First set of results are summarized in Table~\ref{Table:TimeLimitOfSearch}.
Surprisingly, as $T^{search}$ increases, the collision and deadlock rates increase and the success rate decreases except for the change from $\SI{75}{ms}$ to $\SI{150}{ms}$.
This is surprising because, discrete search theoretically produces no worse plans than before as the time allocated for it increases.
The reason of this stems from the dynamic limits of the robot.
We enforce continuity up to acceleration, maximum velocity $\SI{10}{\frac{m}{s}}$ and maximum acceleration $\SI{15}{\frac{m}{s^2}}$.
Enforcing these during trajectory optimization becomes harder as the discrete plan gets more and more complicated, having more rotations and speed changes.
Increasing the time allocated to discrete search makes it more likely to produce complicated plans.
Planning failure rates support this argument.
As $T^{search}$ increases, planning failure rates also increase.

To further support this argument, we run another set of experiments in which we set $c = 1$, $\gamma_1 = \infty$, and $\gamma_2 = \infty$; meaning we enforce continuity up to velocity and do not enforce any derivative magnitude limits.
The results are given in Table~\ref{Table:NoDynTimeLimitOfSearch}.
This time, deadlock and collision rates decrease and the success rate increases as $T^{search}$ increases.
Planning failure rate also increases but its effects are not detrimental to collision and deadlock rates like before.

In the remaining experiments, we set $T^{search} = \SI{75}{ms}$.

\begin{table}[]
    \centering
    \caption{Effects of the time limit of search without dynamic limits to navigation performance.}
    \label{Table:NoDynTimeLimitOfSearch}
    \resizebox{\columnwidth}{!}{\begin{tabular}{|c|c|c|c|c|c|c|}
        \hline \textbf{Exp. $\boldsymbol{\#}$} & \textbf{49} & \textbf{50} & \textbf{51} & \textbf{52} & \textbf{53} & \textbf{54}\\
         \hline $T^{search} [ms]$ & $75$ & $150$ & $300$ & $600$ & $1000$ & $2000$  \\
         \hline 
         \textbf{succ. rate} & 0.828 & 0.892 & 0.922 &  0.934 & 0.926 & 0.933 \\ 
         \textbf{coll. rate} & 0.170 & 0.107 & 0.076 & 0.061 & 0.074 & 0.066\\
         \textbf{deadl. rate} & 0.003 & 0.003 & 0.002 & 0.005 & 0.000 & 0.001\\
         \textbf{s. coll. rate} & 0.003 & 0.000 & 0.000 & 0.000 & 0.000 & 0.000\\
         \textbf{d. coll. rate} & 0.169 & 0.107 & 0.076 & 0.061 & 0.074 & 0.066\\
         \textbf{avg. nav. dur. [s]} & 29.80 & 29.90 & 29.69 & 29.50 & 29.47 & 29.54\\
         \textbf{pl. fail rate} & 0.015 & 0.017 & 0.024 & 0.035 & 0.043 & 0.049\\
         \textbf{avg. pl. dur. [ms]} & 97.27 & 134.80 & 192.80 & 301.26 & 437.43 & 742.61\\
         \hline
    \end{tabular}}
\end{table}

\subsubsection{Desired Planning Horizon}
The desired planning horizon $\tau$ is used by the goal selection stage.
The goal selection stage selects the position on the desired trajectory that is one planning horizon away from the current time as the goal position if robot is collision-free when placed on it.
As $\tau$ increases, our planner tends to plan longer and longer trajectories.

We evaluate the effects of $\tau$ to navigation performance by changing its value and reporting navigation metrics.
In these experiments, we set $\Sigma = \vzero$, $\sigma = 0$, $\rho = 0.2$, and $\#_D = 25$.
We do not mock sensing imperfections of static obstacles.

The results are summarized in Table~\ref{Table:DesiredPlanningHorizon}.
When $\tau$ is set to $\SI{1.0}{s}$, the robot tends to plan shorter trajectories, limiting its longer horizon decision making ability.
Increasing $\tau$ to $\SI{2.5}{s}$, decreases collision and deadlock rates and increases the success rate, as well as the average navigation duration compared to $\tau=\SI{1.0}{s}$.
However, increasing $\tau$ more causes an increase in collision and deadlock rates and a decrease in the success rate.
The reason is that increasing $\tau$ causes distance of goal position to increase in every iteration.
As the distance to the goal position increases, the discrete search step needs more time to produce high quality discrete plans.
Since $T^{search}$ is fixed during these experiments, increasing $\tau$ causes robot to collide or deadlock more.

In the remaining experiments, we set $\tau = \SI{2.5}{s}$.

\begin{table}[]
    \centering
    \caption{Effects of the desired planning horizon to navigation performance.}
    \label{Table:DesiredPlanningHorizon}
    \resizebox{\columnwidth}{!}{\begin{tabular}{|c|c|c|c|c|c|c|}
        \hline \textbf{Exp. $\boldsymbol{\#}$} & \textbf{55} & \textbf{56} & \textbf{57} & \textbf{58} & \textbf{59} & \textbf{60}\\
         \hline $\tau [s]$ & $1.0$ & $2.5$ & $5.0$ & $7.5$ & $10.0$ & $20.0$  \\
         \hline 
         \textbf{succ. rate} & 0.943 & 0.948 & 0.935 & 0.924 & 0.906 & 0.883\\ 
         \textbf{coll. rate} & 0.052 & 0.046 & 0.061 & 0.066 & 0.074 & 0.093 \\
         \textbf{deadl. rate} & 0.008 & 0.006 & 0.004 & 0.011 & 0.022 & 0.031\\
         \textbf{s. coll. rate} & 0.001 & 0.002 & 0.004 & 0.009 & 0.015 &0.049 \\
         \textbf{d. coll. rate} & 0.052 & 0.045 & 0.058 & 0.058 & 0.064 & 0.058\\
         \textbf{avg. nav. dur. [s]} & 29.52 &28.49  & 28.42 & 28.33 & 28.32 &28.61 \\
         \textbf{pl. fail rate} & 0.053 & 0.052 & 0.069 & 0.071 & 0.063 & 0.050\\
         \textbf{avg. pl. dur. [ms]} & 58.71 &  62.55& 80.71 & 119.52 & 152.78 & 195.13\\
         \hline
    \end{tabular}}
\end{table}

\subsubsection{Dynamic Obstacle Sensing Uncertainty}

\begin{table}[]
    \centering
    \caption{Effects of dynamic obstacle sensing uncertainty to navigation performance.}
    \label{Table:DynamicObstacleSensingUncertainty}
    \resizebox{\columnwidth}{!}{\begin{tabular}{|c|c|c|c|c|c|c|}
        \hline \textbf{Exp. $\boldsymbol{\#}$} & \textbf{61} & \textbf{62} & \textbf{63} & \textbf{64} & \textbf{65} & \textbf{66}\\
         \hline $\Sigma [\times I_{2d}]$ & $0.0$ & $0.1$ & $0.1$ & $0.2$ & $0.2$ & $0.5$  \\
         $\sigma$ & $0.0$ & $0.0$ & $0.1$ & $0.1$ & $0.2$ & $0.5$  \\
         \hline 
         \textbf{succ. rate} & 0.905 & 0.879 & 0.832 & 0.797 & 0.661 & 0.410\\ 
         \textbf{coll. rate} &  0.094 & 0.119 &  0.168 & 0.203 & 0.339 &  0.590\\
         \textbf{deadl. rate} &  0.001& 0.003 & 0.002 & 0.001&0.004  & 0.009 \\
         {\color{lightgray}\textbf{s. coll. rate}} & {\color{lightgray}0.000} & {\color{lightgray}0.000} & {\color{lightgray}0.000} & {\color{lightgray}0.000} & {\color{lightgray}0.000}  &  {\color{lightgray}0.000}\\
         \textbf{d. coll. rate}& 0.094 & 0.119 & 0.168 & 0.203& 0.339 &  0.590\\
         \textbf{avg. nav. dur. [s]} & 26.71 & 26.70 & 26.70 &26.71 &26.69  &  26.72\\
         \textbf{pl. fail rate}& 0.026 & 0.030 & 0.028 & 0.029& 0.031 & 0.037 \\
         \textbf{avg. pl. dur. [ms]}&  59.76& 55.53 & 54.54 & 54.83& 54.46 & 56.15 \\
         \hline
    \end{tabular}}
\end{table}

\begin{table}[]
    \centering
    \caption{Effects of dynamic obstacle inflation to navigation performance under dynamic obstacle sensing uncertainty of $\Sigma = 0.2I_{2d}, \sigma = 0.2$.}
    \label{Table:InflationUnderDynamicObstacleSensingUncertainty}
    \resizebox{\columnwidth}{!}{\begin{tabular}{|c|c|c|c|c|c|c|}
        \hline \textbf{Exp. $\boldsymbol{\#}$} & \textbf{67} & \textbf{68} & \textbf{69} & \textbf{70} & \textbf{71} & \textbf{72}\\
         \hline inflation $[m]$ & $0.2$ & $0.5$ & $1.0$ & $1.5$ & $2.0$ & $4.0$  \\
         \hline 
         \textbf{succ. rate} & 0.779 & 0.818 & 0.747 & 0.617 & 0.417 & 0.057\\ 
         \textbf{coll. rate} & 0.220 & 0.182 & 0.250 & 0.382 & 0.581 &  0.943\\
         \textbf{deadl. rate}& 0.003 & 0.001 &  0.005& 0.007 & 0.003 &  0.000\\
         \textbf{s. coll. rate}& 0.000 & 0.000 & 0.000 & 0.000  & 0.000 &  0.000\\
         \textbf{d. coll. rate} & 0.22 & 0.182 & 0.250 & 0.382 & 0.581 &  0.943\\
         \textbf{avg. nav. dur. [s]} & 26.70 & 26.72 & 26.93 & 27.16 & 27.79 & 26.81 \\
         \textbf{pl. fail rate} & 0.032 & 0.035 & 0.047 & 0.058 &0.067 &  0.021\\
         \textbf{avg. pl. dur. [ms]} & 57.80 &  63.014& 70.92 & 74.63 & 72.66 & 23.13 \\
         \hline
    \end{tabular}}
\end{table}

The dynamic obstacle sensing uncertainty is mocked by i) applying a zero mean Guassian with covariance $\Sigma$ to the sensed positions and velocities of dynamic obstacles and ii) randomly inflating dynamic obstacle shapes $\mR_\fD$ by a zero mean Gaussian with standard deviation $\sigma$.
These create three inaccuracies reflected to our planner as mentioned in Section~\ref{Section:Mocking}: i) dynamic obstacle shapes fed to our planner become wrong, ii) prediction inaccuracy increases, and iii) current positions $\vp^{current}_\fD$ of dynamic obstacles fed to the planner become wrong.
Note that our planner does not explicitly model dynamic obstacle shape sensing uncertainty.
In addition, it does not explicitly model dynamic obstacle current position uncertainty.
It also does not explicitly model uncertainty of predicted individual behavior models.
It models uncertainty across behavior models by assigning probabilities to each of them.

To evaluate our planner's performance under different levels of dynamic obstacle sensing uncertainty, we control $\Sigma$ and $\sigma$ and report the metrics.

During these experiments, we set $\rho = 0$ to evaluate the effects of dynamic obstacles only. 
We set $\#_D = 50$.
The results are given in Table~\ref{Table:DynamicObstacleSensingUncertainty}.
$\Sigma$ is set to a constant multiple of identity matrix $I_{2d}$ of size $2d \times 2d$ in each experiments.
Expectedly, as the uncertainty increases, success rate decreases.
Increase in collision and deadlock rates are also seen.
Similarly, planning failure rate tends to increase as well.

One common approach to tackle unmodeled uncertainty for obstacle avoidance is artificially inflating shapes of obstacles.
To show the effectiveness of this approach, we set $\Sigma = 0.2I_{2d}$ and $\sigma = 0.2$, and inflate the shapes of obstacles with different amounts during planning.
The results of these experiments are given in Table~\ref{Table:InflationUnderDynamicObstacleSensingUncertainty}.
The inflation amount is the amount we inflate the shape of the dynamic obstacle in all axes (e.g., if inflation amount is $\SI{0.2}{m}$, obstacle's size is increased by $\SI{0.2}{m}$ in $x$, $y$, and $z$ axes simultaneously).
Inflation clearly helps when done in reasonable amounts.
When inflation is set to $\SI{0.2}{m}$, success rate increases from $0.661$ as reported in Table~\ref{Table:DynamicObstacleSensingUncertainty} to $0.779$.
It further increases to to $0.818$ when inflation amount is set to $\SI{0.5}{m}$.
However, as the inflation amount increases more, metrics start to degrade as the planner becomes overly conservative.
Success rate decreases down to $0.057$ when the inflation amount is set to $\SI{4.0}{m}$.

In the remaining experiments, we set $\Sigma = \vzero$ and $\sigma = 0$.

\subsubsection{Static Obstacle Sensing Uncertainty}

\begin{table*}[]
    \centering
    \caption{Effects of static obstacle sensing uncertainty to navigation performance.}
    \label{Table:StaticObstacleSensingUncertainty}
    \begin{tabular}{|c|c|c|c|c|c|c|}
        \hline \textbf{Exp. $\boldsymbol{\#}$} & \textbf{73} & \textbf{74} & \textbf{75} & \textbf{76} & \textbf{77} & \textbf{78}\\
         \hline mocking & $L(0.2)$ & $L(0.2)^2$ & $L(0.3)^2I$ &$L(0.3)^2ID$& $L(0.3)^2IDL(0.2)$ & $L(0.3)^2IDL(0.5)$   \\
         \hline 
         \textbf{succ. rate} & 0.994 & 0.992 & 0.986 & 0.956 & 0.971 & 0.949 \\
         \textbf{coll. rate} &0.001  & 0.003&0.005 &  0.037&0.020 &0.039\\
         \textbf{deadl. rate}&0.005 &0.005 &0.009 & 0.007& 0.009 & 0.015\\
         \textbf{s. coll. rate}&0.001 &0.003 & 0.005& 0.037&0.020 &0.039 \\
         {\color{lightgray}\textbf{d. coll. rate}}&{\color{lightgray}0.000} & {\color{lightgray}0.000}& {\color{lightgray}0.000}&{\color{lightgray}0.000} & {\color{lightgray}0.000}&{\color{lightgray}0.000} \\
         \textbf{avg. nav. dur. [s]}&27.49 & 27.53& 27.73&27.49 & 27.55& 28.18\\
         \textbf{pl. fail rate}&0.029 &0.041 & 0.053& 0.038 & 0.047 & 0.063\\
         \textbf{avg. pl. dur. [ms]}& 36.16& 41.76&53.39 &40.58 & 46.41 &67.65 \\
         \hline
    \end{tabular}
\end{table*}

As we describe in Section~\ref{Section:Mocking}, we use $3$ operations to mock sensing imperfections of static obstacles: i) increaseUncertainty, ii) leakObstacles($p_{leak}$), and iii) deleteObstacles.
We evaluate the effects of static obstacle sensing uncertainty by applying a sequence of these operations to the octree representation of static obstacles, and provide the resulting octree to our planner.
Application of leakObstacles increases the density of the map, but the resulting map contains the original obstacles. 
increaseUncertainty does not change the density of the map, but increases the uncertainty associated with the obstacles.
deleteObstacles decreases the density of the map, but the resulting map may not contain the original obstacles, leading to unsafe behavior.

During these experiments, we set $\rho = 0.1$.
We set $\#_D = 0$ to evaluate the effects of static obstacles only.

The results are given in Table~\ref{Table:StaticObstacleSensingUncertainty}.
In mocking row of the table, we use $L(p_{leak})$ as an abbreviation for leakObstacles($p_{leak}$), and $L(p_{leak})^n$ as an abbreviation for repeated application of leakObstacles to the octree.
We use $I$ for increaseIncertainty, and $D$ for deleteObstacles.
Leaking obstacles or increasing the uncertainty associated with them does not increase the collision and deadlock rates significantly as seen in experiments $73, 74,$ and $75$.
The planning duration and failure rates increase as the number of obstacles in the environment increases.
Deleting obstacles causes a sudden jump of the collision rate from experiment $75$ to $76$ because the planner simply does not know about existing obstacles.
Leaking obstacles back with $p_{leak} = 0.2$ after deleting them decreases the collision rate back from experiment $76$ to $77$.
However, leaking obstacles with high probability increases the collision rate back from experiment $77$ to $78$.
This happens because the environments get significantly more complicated because the number of obstacles increase.
Complication of the environments can also be seen by the increased deadlock rate.

\subsection{Comparison with a Baseline}

We compare our algorithm with MADER~\cite{tordesillas2020mader}, which is a real-time trajectory planning algorithm for static and dynamic obstacles avoidance as well as asynchronous multi-robot collision avoidance.
During our comparison, we focus on static and dynamic obstacle avoidance capabilities of MADER.

MADER models dynamic obstacle movements using predicted trajectories.
It does not explicitly model ego robot--dynamic obstacle interactions.
It supports uncertainty associated with predicted trajectories using axis aligned boxes, which we call uncertainty boxes from now on, such that it requires that the samples of the real trajectory dynamic obstacle is going to follow is contained in known bounding boxes around the samples of the predicted trajectory.

Dynamic obstacles move according to movement and interaction models in our formulation.
Converting movement models to predicted trajectories is possible by propagating dynamic obstacles' positions according to the desired velocities returned from the movement models.
However, since the interactive behavior of dynamic obstacles depend on the trajectory that the ego robot is going to follow, which is computed by the planner itself, their effect to the future trajectories are unknown prior to planning.
Therefore, we convert predicted behavior models of dynamic obstacles to predicted trajectories by propagating dynamic obstacle positions with movement models, and model interactive behavior as uncertainty by setting uncertainty boxes during evaluation.

When a dynamic obstacle is not interactive, e.g. repulsion strength is $0$ in a repulsive model, predicted trajectories are perfect if the decision making period of dynamic obstacles are known and movement model predictions are correct.

MADER enforces dynamic limits independently for each dimension similar to our planner.
It creates outer polyhedral representations of dynamic obstacles' shapes inflated by the uncertainty boxes around the predicted trajectories, and avoids them during planning.
Planning pipeline of MADER consists of a discrete search method followed by trajectory optimization similar to ours.
The optimization problem is a non-linear program that is locally optimized using the discrete solution as initial guess.
It plans trajectories toward given goal positions.
We add support of desired trajectories to MADER by running our goal selection algorithm in every planning iteration, and provide the selected goal position to MADER.
During our comparisons, we use \emph{with prior} strategy for both MADER and our algorithm that the desired trajectories avoid the static obstacles in the environment.

During our comparison, we use the code of MADER published by its authors\footnote{\url{https://github.com/mit-acl/mader}} and integrate it to our simulation system.
In MADER's implementation, the ego robot is modeled as a sphere, while the algorithm can support any convex collision shape in theory.
We implement our algorithm for box like collision shapes.
To provide robot shapes we have to MADER, we compute the smallest spheres containing the boxes.
In order not to cause MADER to be overly conservative, we change our robot shape randomization method to generate boxes that have the same sizes in all dimensions.
We also sample robot sizes in interval $[\SI{0.1}{m}, \SI{0.2}{m}]$ instead of $[\SI{0.2}{m}, \SI{0.3}{m}]$ so that the robot can fit between static obstacles easily when its collision shape is modelled using bounding spheres. (Since resolution of octrees we generate is $\SI{0.5}{m}$, the smallest possible gap between static obstacles is $\SI{0.5}{m}$.) 
The randomization of our runs are same as before except for these changes.
We allow MADER to run up to $\SI{500}{ms}$ in each planning iteration even when the simulated replanning period is smaller.
We freeze the environment until MADER is done replanning to cancel the affects of exceeding replanning period.
Having lower limits on planning time, e.g. $\SI{350}{ms}$, decreases the performance of MADER drastically ($\approx50\%$ more collision rate) in our experiments.
Instead of generating $1000$ random environments for each experiment, we generate $250$ random environments while evaluating MADER because the MADER simulations are considerably slower than ours.

\begin{table}[]
    \centering
    \caption{Performance of MADER when it avoids most likely or all behavior models of dynamic obstacles.}
    \label{Table:MADERAvoidAllOrNot}
    \resizebox{\columnwidth}{!}{\begin{tabular}{|c|c|c|c|c|c|c|}
        \hline \textbf{Exp. $\boldsymbol{\#}$} & \textbf{79} & \textbf{80} & \textbf{81} & \textbf{82} & \textbf{83} & \textbf{84}\\
         \hline avoid & all & most l. & all & most l. & all & most l.  \\
         $\rho$ & 0.1 & 0.1 & 0.2& 0.2 &0.2&0.2\\
         $\#_D$ & 10 & 10 & 10& 10 & 25 & 25  \\
         \hline 
         \textbf{succ. rate} & 0.828 & 0.848 & 0.640 & 0.672 &0.352 & 0.380 \\
         \textbf{coll. rate} & 0.160 & 0.112 & 0.200& 0.200 &0.472 & 0.520  \\
         \textbf{deadl. rate} & 0.076 & 0.092 &0.276 & 0.256 &0.580 & 0.504 \\
         \textbf{s. coll. rate} & 0.000 & 0.000 & 0.000& 0.000 &0.000 & 0.000  \\
         \textbf{d. coll. rate} & 0.160 & 0.112 & 0.200 & 0.200 &0.472 & 0.520  \\
         \textbf{avg. nav. dur. [s]} & 24.85 & 24.87 & 25.94& 25.99 &25.93 & 26.05 \\
         \textbf{pl. fail rate} &  0.248 & 0.255 & 0.514& 0.486 &0.739 &  0.702  \\
         \textbf{avg. pl. dur. [ms]}& 266.97 & 260.26 &398.53 & 387.06 &426.41 & 407.47  \\
         \hline
    \end{tabular}}
\end{table}

\subsubsection{Avoiding All or Most Likely Behavior Models of Dynamic Obstacles with MADER}\label{Section:MADERAvoidAllOrNot}
Our prediction system generates $3$ behavior models for each dynamic obstacle and assigns probabilities to them.
MADER does not support multiple behavior hypothesis for each obstacle explicitly.
Therefore, to avoid a dynamic obstacle, we have to choice of avoiding its most likely behavior model or all of its behavior models modelling each behavior model as a separate obstacle.
Another option is to model the dynamic obstacle as a single obstacle and set the uncertainty box that it contains all predicted trajectories generated from each behavior model.
This is an extremely conservative approach  because the predictions can be pointing to completely different directions in completely different speeds, because of which we do not evaluate this possibility.

To evaluate the performance of MADER between these two choices, we run it in environments with different densities and compare the approaches.
During these experiments, we set repulsion strength $f = 0$.
In addition, decision making period of dynamic obstacles are perfectly known, meaning that predicted trajectories are correct if predicted movement models are correct.
We set the uncertainty box to a symmetric one with size $\SI{0.1}{m}$ in all dimensions to handle prediction inaccuracy of individual movement models.

The results are summarized in Table~\ref{Table:MADERAvoidAllOrNot} for varying static obstacle density $\rho$ and number of dynamic obstacles $\#_D$.
The success rates increase when MADER avoids only the most likely behavior models of dynamic obstacles compared to avoiding all behavior models.
Surprisingly, between experiments $79$ and $80$, collision rate decreases even if the algorithm becomes more conservative in terms of safety by avoiding all modes.
The success, collision, and deadlock rates in experiments $83$ and $84$ suggests that deadlocks and collisions occur in the same runs a lot.
For instance, in experiment $84$, success rate is $0.380$, collision rate is $0.520$, and the deadlock rate is $0.504$, suggesting that $\approx 40\%$ of cases are cases in which robot deadlocks and collides, accounting for $\approx 78\%$ collision cases.

\begin{table}[]
    \centering
    \caption{Performance of MADER when different sizes of prediction uncertainty boxes are used.}
    \label{Table:MADERUncertaintyBox}
    \resizebox{\columnwidth}{!}{\begin{tabular}{|c|c|c|c|c|c|c|}
        \hline \textbf{Exp. $\boldsymbol{\#}$} & \textbf{85} & \textbf{86} & \textbf{87} & \textbf{88} & \textbf{89} & \textbf{90}\\
         \hline uncer. b. size $[m]$ & $0.0$ &$0.2$& $0.5$ & $1.0$ & $2.0$ & $3.0$   \\
         \hline 
         \textbf{succ. rate} & 0.736 & 0.724 & 0.672 & 0.616 & 0.468 & 0.220\\
         \textbf{coll. rate} & 0.248 & 0.236 & 0.272 & 0.276 & 0.268 & 0.296\\
         \textbf{deadl. rate} & 0.124  & 0.152 & 0.188 & 0.284 & 0.452 & 0.724\\
         \textbf{s. coll. rate} & 0.000 & 0.000 & 0.000  & 0.000 & 0.000 & 0.000 \\
         \textbf{d. coll. rate} & 0.248 & 0.236 & 0.272  & 0.276 & 0.268 & 0.296\\
         \textbf{avg. nav. dur. [s]} & 25.73 & 25.85 &26.04  & 26.99 & 28.38 & 30.01\\
         \textbf{pl. fail rate} & 0.342 & 0.400 &  0.459 & 0.570 &  0.700 & 0.828\\
         \textbf{avg. pl. dur. [ms]} & 388.50 & 384.76 & 390.22 & 390.69 & 383.81 & 395.57 \\
         \hline
    \end{tabular}}
\end{table}

\begin{table*}[]
    \centering
    \caption{Comparison of MADER with our planner}
    \label{Table:MADERVSOurs}
    \resizebox{\linewidth}{!}{\begin{tabular}{|c|c|c|c|c|c|c|c|c|c|c|c|}
        \hline \textbf{Exp. $\boldsymbol{\#}$} & $\rho$ & $\#_D$ & Alg. & \textbf{succ. rate} & \textbf{coll. rate} & \textbf{deadl. rate} & \textbf{s. coll. rate} & \textbf{d. coll. rate} & \textbf{avg. nav. dur. [s]} & \textbf{pl. fail rate} & \textbf{avg. pl. dur. [ms]}\\
        \hline \multirow{2}{*}{\textbf{91}} & \multirow{2}{*}{$0.0$} & \multirow{2}{*}{$15$} & MADER & 0.980 & 0.020 & 0.008 & \textbf{0.000} & 0.020 &\textbf{24.10}&0.051&32.26\\
        &&&Ours & \textbf{0.996} & \textbf{0.004} &\textbf{0.000}& \textbf{0.000}& \textbf{0.004}& 26.70& \textbf{0.009}& \textbf{21.05}\\
        \hline \multirow{2}{*}{\textbf{92}} & \multirow{2}{*}{$0.1$} & \multirow{2}{*}{$15$} & MADER & 0.820 & 0.180 & 0.028 & \textbf{0.000} & 0.180 & \textbf{24.96} & 0.170 & 264.11\\
        &&&Ours &\textbf{0.980}&\textbf{0.016}&\textbf{0.004}&\textbf{0.000}&\textbf{0.016}&27.41&\textbf{0.032}&\textbf{42.33}\\
        \hline \multirow{2}{*}{\textbf{93}} & \multirow{2}{*}{$0.2$} & \multirow{2}{*}{$15$} & MADER &0.724&0.248&0.116&\textbf{0.000}&0.248&\textbf{25.86}&0.352&393.28\\
        &&&Ours &\textbf{0.976}&\textbf{0.020}&\textbf{0.004}&0.004&\textbf{0.016}&28.28&\textbf{0.053}&\textbf{49.79}\\
        \hline \multirow{2}{*}{\textbf{94}} & \multirow{2}{*}{$0.2$} & \multirow{2}{*}{$25$} & MADER &0.616&0.364&0.192&\textbf{0.000}&0.364&\textbf{25.76}&0.465&401.95\\
        &&&Ours &\textbf{0.960}&\textbf{0.040}&\textbf{0.000}&\textbf{0.000}&\textbf{0.040}&28.24&\textbf{0.050}&\textbf{57.32}\\
        \hline \multirow{2}{*}{\textbf{95}} & \multirow{2}{*}{$0.2$} & \multirow{2}{*}{$50$} & MADER &0.292&0.664&0.484&\textbf{0.000}&0.664&\textbf{25.82}&0.711&423.01\\
        &&&Ours &\textbf{0.884}&\textbf{0.116}&\textbf{0.000}&0.008&\textbf{0.116}&28.32&\textbf{0.058}&\textbf{75.39}\\
        \hline \multirow{2}{*}{\textbf{96}} & \multirow{2}{*}{$0.3$} & \multirow{2}{*}{$50$} & MADER &0.228&0.732&0.620&\textbf{0.000}&0.732&\textbf{27.32}&0.776&491.27\\
        &&&Ours &\textbf{0.868}&\textbf{0.124}&\textbf{0.016}&0.020&\textbf{0.120}&29.81&\textbf{0.079}&\textbf{86.28}\\
        \hline
    \end{tabular}}
\end{table*}

In the remaining experiments, we provide the predictions generated from most likely behavior models of dynamic obstacles to MADER because it results in a better performance.

\subsubsection{Setting Uncertainty Box with MADER}\label{Section:MADERUncertaintyBox}

Next, we simulate the behavior of MADER with interactive obstacles, and evaluate the performance of it when uncertainty boxes with different sizes are used.

We set $\rho = 0.2$, and $\#_D = 15$ during these experiments.
Repulsion strength $f$ is sampled uniformly in interval $[0.2, 0.5]$ and the dynamic obstacle decision making period is sampled uniformly in interval $[\SI{0.1}{s}, \SI{0.5}{s}]$.
These create prediction inaccuracies for MADER stemming from $3$ main reasons: i) individual movement models are inaccurate because the decision making periods are unknown, ii) movement models do not explain the behavior of the dynamic obstacles because of interactivity, and iii) predicted behavior models are inaccurate because of the inherent inaccuracy of predictors listed in Section~\ref{Section:Prediction}.

We set uncertainty boxes to axis aligned boxes with sizes from $\SI{0.0}{m}$ to $\SI{3.0}{m}$ in each axis. 
The results are given in ~\ref{Table:MADERUncertaintyBox}.
The success rates decrease as the size of the uncertainty box increases.
The increase in collision rate is smaller compared to the increase in deadlock rate.
As the uncertainty box size increases, avoiding dynamic obstacles become harder, causing robot to divert from the desired trajectory more.
As the robot diverts from the desired trajectory, the likelihood that it deadlocks because of static obstacles increases.
Therefore, even if the uncertainty box handles uncertainty associated with dynamic obstacle predictions, the fact that it drives robots away from the desired trajectory increases its deadlock rate. 
+

In the remaining experiments, we set the uncertainty box size to $\SI{0.0}{m}$ for MADER because it results in a better performance.

\subsubsection{Comparison of Our Planner and MADER}

We compare our planner with MADER in environments with different static obstacle densities $\rho$ and number of dynamic obstacles $\#_D$.

During these experiments, the repulsion strength $f$ of dynamic obstacles is sampled in interval $[0.2, 0.5]$ uniformly, and the decision making period of dynamic obstacles are sampled in interval $[\SI{0.1}{s}, \SI{0.5}{s}]$ uniformly.

The results are summarized in Table~\ref{Table:MADERVSOurs}.
In general, our algorithm results in a considerably higher success rate, lower collision, deadlock, planning failure rates, and a lower average planning duration.
MADER results in a lower navigation duration compared to our planner.
We attribute the reason of this to our planner's behavior during the tail end of the navigation.
When the robot is close to the goal position, the planning horizon $\tau'$ of the search is set to the minimum search horizon parameter $\tilde{\tau}$, which is set to $\SI{2.0}{s}$ in our experiments. 
Therefore, the search always aims to reach to the goal position within at least $\SI{2.0}{s}$, causing the robot to slow down during the tail end of the navigation.
MADER results in $0$ static obstacle collision rates in all cases, because it never generates a trajectory that collides with static obstacles.
Since it also plans for stopping at the end of the trajectory, robot is either following a safe trajectory with respect to static obstacles or is stopped.
On the other hand, our planner occasionally results in collisions with static obstacles because we have no such property.

\subsection{Physical Robot Experiments}

We implement and run planner for Crazyflie 2.1s.
During physical experiments, we use $6$ Crazyflie 2.1s as dynamic obstacles moving according to goal attractive, rotating, or constant velocity movement models and repulsive interaction model.
A planning Crazyflie 2.1, i.e. the ego robot, runs the predictors we describe in real-time to generate a probability distribution over behavior models of each dynamic obstacle.
Then, it runs our planner to compute trajectories avoiding the dynamic obstacles in real-time.

For dynamic obstacle and ego robot localization, we use VICON motion tracking system, and we manage the Crazyflies using Crazyswarm~\cite{preiss2017crazyswarm}.
We use Robot Operating System (ROS)~\cite{quigley09ros} as the underlying software system.
Predictors and our algorithm run in a basestation computer with Intel(R) Xeon(R) CPU E5-2630 v4 @2.20GHz CPU, running Ubuntu 20 as the operating system.

Our physical experiments show the feasibility of running our approach in real-time in the real world.
The recordings from our physical experiments can be found in our accompanying video at \url{https://youtu.be/0XAvpK3qX18}.

\section{Conclusion}
We present a probabilistic real-time trajectory planning algorithm for mobile robots navigating in environments with static and interactive dynamic obstacles.
The planner explicitly minimizes collision probabilities against static and dynamic obstacles as well as distance, duration, and rotations using a multi-objective search method; and energy usage during optimization.
The behavior of dynamic obstacles are modeled using two vector fields, namely movement and interaction models, in velocity space.
Movement models describe the intentions of the dynamic obstacles, while the interaction models describe the inteaction of dynamic obstacles with the ego robot.
The planner simulates the decisions of dynamic obstacles during decision making in response to the actions it is going to take using the interaction model.
We present $3$ online model based prediction algorithms to predict the behavior models of dynamic obstacles and assign probabilities to them.

We evaluate our algorithm extensively in different environments and configurations using a total of $78000$ randomly generated environments.
We compare our algorithm with a state-of-the-art real-time trajectory planning algorithm for static and dynamic obstacle avoidance in $4500$ randomly generated environments, and show that our planner achieves a higher success rate.
We show our algorithm's feasibility in the real-world by implementing and running it for Crazyflie 2.1s.

Future work includes integrating cooperative multi-robot collision avoidance to our planner as well as generalizing the search stage to other state/action spaces.

\bibliographystyle{IEEEtran}
\bibliography{bibliography}

\vfill

\end{document}